\documentclass[10pt, a4paper]{article}

\usepackage[final]{lrec2026} 

\usepackage{subcaption}
\usepackage{fdsymbol}

\newcommand{\lc}[1]{\textcolor{black}{#1}}

\title{Voice, Bias, and Coreference: \\ An Interpretability Study of Gender in Speech Translation}

\name{
    Lina Conti$^{\spadesuit\vardiamondsuit}$,
    Dennis Fucci$^{\spadesuit\vardiamondsuit}$,
    Marco Gaido$^\spadesuit$,
    Matteo Negri$^\spadesuit$,
\\
    {\large \bf Guillaume Wisniewski}$^\clubsuit$,
    {\large \bf Luisa Bentivogli}$^\spadesuit$
\\}

\address{$^\spadesuit$Fondazione Bruno Kessler, Italy \\ 
          \{lvarellaconti,dfucci,mgaido,negri,bentivo\}@fbk.eu \\ \\
          $^\vardiamondsuit$University of Trento, Italy \\ \\
          $^\clubsuit$Laboratoire de Linguistique Formelle, Université Paris Cité, CNRS, Paris, France \\ 
          guillaume.wisniewski@u-paris.fr \\}

\abstract{
Unlike text, speech conveys information about the speaker, such as gender, through acoustic cues like pitch. This gives rise to modality-specific bias concerns.
For example, in speech translation (ST), when translating from languages with notional gender, such as English, into languages where gender-ambiguous terms referring to the speaker are assigned grammatical gender, the speaker’s vocal characteristics may play a role in gender assignment.
This risks misgendering speakers—whether through masculine defaults or vocal-based assumptions—yet how ST models make these decisions remains poorly understood.
We investigate the mechanisms ST models use to assign gender to speaker-referring terms across three language pairs (en$\rightarrow$es/fr/it).
\lc{To do so, we examine} how training data patterns, internal language model (ILM) biases, and acoustic information interact. We find that models do not simply replicate term-specific gender associations from training data, but learn broader patterns of masculine prevalence. While the ILM exhibits strong masculine bias, models can override these preferences based on acoustic input. Using contrastive feature attribution on spectrograms, we reveal that the model with higher gender accuracy relies on a previously unknown mechanism: using first-person pronouns to link gendered terms back to the speaker, accessing gender information distributed across the frequency spectrum rather than concentrated in pitch.
 \\ \newline \Keywords{gender bias, speech translation, interpretability, XAI} }

\begin{document}

\maketitleabstract

\section{Introduction}

Improved speech technology has made voice a popular modality to interact with AI systems, with applications like live translation through earphones now widely available \cite{chen-2025-airpods}.
Unlike text, speech conveys information beyond linguistic content: vocal characteristics like pitch and pronunciation provide cues about the speaker's sociodemographic attributes, including gender, age, race, and social class \cite{labov1964phonological, Thomas2010TeachingAL, kraus2017signs, simpson2009phonetic}.
This raises modality-specific concerns about social bias, as systems may perform differently across groups defined by these acoustic characteristics \cite{doi:10.1073/pnas.1915768117, curry-etal-2024-classist}. 

One example of these modality-specific concerns is the manifestation of gender bias when translating speech input compared to text input.
When translating from languages with limited gender marking like English to languages with overt gender distinctions like Spanish, French, and Italian, systems assign grammatical gender to ambiguous terms. 
For example, when translating “\textit{I have become a student}” to Italian, the verb form for “\textit{become}” is grammatically gendered, leading the model to choose between “\textit{diventata}”$^F$ (feminine) and “\textit{diventato}”$^M$ (masculine).
In text-based machine translation (MT), systems typically default to masculine forms or make assumptions based on gender stereotypes \cite{Prates2018AssessingGBA, stanovsky-etal-2019-evaluating, mastromichalakis2025assumed}. 
Speech translation (ST) systems can exhibit similar patterns, but they also have access to vocal characteristics, such as pitch, that could be used as proxies for the speaker's gender when translating terms that refer to them \cite{bentivogli-etal-2020-gender}, as in the example above.

However, whether and how ST models use acoustic information for gender assignment remains poorly understood. While interpretability methods have been used to better understand gender bias in MT \cite{vanmassenhove2019lost, wisniewski2022analyzing, attanasio-etal-2023-tale, manna-etal-2025-paying}, research on the mechanisms underlying gender assignment in ST is scarce \cite{xu2023recent, fucci-etal-2025-different, yang2025towards}. This gap is critical: without understanding how models make these ethically sensitive decisions, developing targeted mitigation for gender bias becomes significantly more challenging.

To investigate the mechanisms ST models use to assign gender to speaker-referring terms, we start from the common assumption that attributes gender bias to training data imbalances \cite{tatman2017gender, garnerin2019gender, iluz-etal-2023-exploring, mastromichalakis2025assumed}. This leads to our first research question: (i) \textbf{What is the influence of gender associations learned from the training data?} We address this by comparing model predictions with gender frequencies in the training corpus (\S \ref{sec:training-data}). Finding that models do not simply replicate term-specific patterns motivates us to investigate the broader factors driving gender assignment in ST models.  
For this, we conceptually divide the ST model into two components: the encoder, which processes the input audio and may extract acoustic cues from it, and the decoder, which autoregressively predicts the next token based on both the encoder's representation of the audio and the previously generated tokens.
First, we study the decoder's contribution by removing encoder information, thus isolating the ST system's internal language model (ILM) \cite{variani2020hybrid,meng2021internal,zeineldeen2021investigating}. Through this, we aim to answer the question (ii) \textbf{What is the impact of the model's learned knowledge of the target language and a priori assumptions about gender on predictions?} Following this analysis (\S \ref{sec:ilm}), we turn to assessing the role of the source audio: (iii) \textbf{What aspects of the input audio does the model use to assign gender to speaker-referring terms?} Does it rely primarily on pitch, a key acoustic correlate of perceived gender? We study this 
\lc{through} contrastive feature attribution over input spectrograms (\S \ref{sec:feat-attribution}).

The findings of our analysis across three language pairs (en$\rightarrow$es/fr/it) challenge common assumptions about how ST systems perform gender assignment. The models we study do not simply replicate term-specific associations from training data\lc{,} but learn broader patterns of masculine prevalence. While the ILM exhibits masculine bias, models can override these preferences. Moreover, they use first-person pronouns to link the gendered term back to the speaker, accessing vocal cues 
\lc{distributed across the frequency spectrum. This challenges the assumption that pitch would be the key acoustic feature \cite{bentivogli-etal-2020-gender,fucci2023no}, as we find the first and second formants to be more important.}

\section{Bias Statement}
\label{sec:bias-statement}

Following \citet{blodgett-etal-2020-language}, we make explicit the  assumptions underlying our work on bias. We focus on misgendering in ST: when systems translate 
speaker-referring terms into gendered target language forms that do not align with the speaker's gender identity. We consider outputs biased when they contradict reference translations that reflect 
\lc{the gender the speaker identifies with.}
When ST systems misgender speakers, allocational harms can arise through unequal performance: misgendered users must spend resources correcting system outputs \cite{savoldi-etal-2024-harm}. It also creates representational harms through the invisibilization of 
genders \lc{other than masculine}, since models typically default to masculine forms. These harms affect women and gender non-conforming individuals. The binary gender framework we follow in this analysis does not allow us to cover the latter group; we discuss this limitation in \S\ref{sec:ethics-statement}.

\section{Related Works}

Gender bias in MT has been extensively studied \cite{savoldi2025decade} with interpretability work revealing mechanisms underlying gendered choices in text-based systems.
For instance, \citet{wisniewski2022analyzing} and \citet{manna-etal-2025-paying} show that accurate gender disambiguation critically depends on correct handling of coreference chains.
Feature attribution analyses \cite{attanasio-etal-2023-tale, sarti2023inseq} further reveal that incorrect predictions typically result from models failing to attend to coreferring gendered pronouns and following a general masculine default rather than term-specific stereotypes.
These insights from MT motivate our application of similar interpretability methods to ST, particularly given their success in informing mitigation strategies \cite{attanasio-etal-2023-tale, sarti2023inseq}.

However, ST introduces a distinct dimension that requires specific consideration. 
The same sentence sounds different depending on the speaker's gender, with acoustic variations in pitch, resonant frequencies, voice quality, and speech rate arising from both biological factors and sociocultural learned patterns \cite{simpson2009phonetic, kreiman2011foundations, azul2015varied}. Both humans and machines can detect and react to these variations \cite{tusing2000sounds, chao2021girl, brown2025sociophonetic}.
\lc{Studying whether and how ST models leverage these acoustic cues requires methods specifically designed for the speech modality, beyond those developed for gender bias in text-based systems.}

Existing work on gender bias in speech technology has primarily focused on documenting performance disparities across demographic groups in various tasks: emotion recognition \cite{slaughter-etal-2023-pre, chien2024balancing, lin2024emo}, automatic speech recognition (ASR) \cite{adda2005speech, sawalha2013effects, tatman2017gender, garnerin2019gender, feng2021quantifying, liu2022towards, meng2022don, rajan2022aequevox, attanasio-etal-2024-twists}, and ST \cite{zanon2022study, costa-jussa-etal-2022-evaluating, bansal2025addressing}. 
Another common line of work studies the gender bias in speech technologies arising from what is said rather than how it sounds.
Many studies adapt textual bias benchmarks by generating audio versions through text-to-speech systems \cite{lin2024spokenstereoset, lin2024listen}, which primarily test content-triggered bias. 
\lc{While this body of work establishes that \lc{gender} bias 
\lc{is present in speech}, it does not explain the mechanisms through which models use acoustic information.}

Some recent studies have begun examining bias related to acoustic gender cues in spoken question answering \cite{choi2025voicebbq, choi2025acoustic}, finding that models largely fail to use them effectively and appropriately.
But the translation task differs from question answering, and the pressure to assign grammatical gender could incentivize ST models to extract and use acoustic information differently.

Beyond measuring bias, some work has investigated the mechanisms behind gender bias in ST. \citet{savoldi2022dynamics} examined how it emerges during training, and \citet{savoldi-etal-2022-morphosyntactic} studied how tokenization choices affect gender bias patterns. However, these works do not explore how models use acoustic information for gender assignment, which is the focus of this paper.

Interpretability research has shown that speech representations encode gender information \cite{de2022probing, chowdhury2024end, guillaume2024gender, krishnan2024encoding, lin2024social, fucci-etal-2025-different}.
However, these studies typically fail to establish causal links between encoded information and model outputs. 
We address this limitation by using perturbation-based feature attribution (\S\ref{method:feat-attribution}) to identify features that causally drive gender assignment.

\section{Method}

To investigate the factors driving gender assignment by ST models for terms referring to the speaker, we examine three potential sources: training data patterns, the decoder's learned biases independent of the input audio, and the most relevant acoustic features from the input for gender assignment.
In this section, we introduce the methods through which we investigate these aspects: comparison of the model's prediction with frequency patterns in the training data (§\ref{method:training-data}); ILM analysis to examine the decoder's learned biases (§\ref{method:ilm}); and contrastive feature attribution to identify the aspects of the audio driving gender assignment (§\ref{method:feat-attribution}).

\subsection{Training Data Prevalence}
\label{method:training-data}

To test the common assumption that gender bias is merely a direct reflection of training data distribution,
we examine whether the model's gender preferences align with gender prevalence in its training data. The models we analyze \cite{wang2020fairseq,papi2024good} are trained on a single open-source dataset, MuST-C \cite{cattoni2021must}, which enables this analysis.

We identify gender terms referring to the speaker through string matching with the gender annotations in MuST-SHE \cite{bentivogli-etal-2020-gender}, our evaluation benchmark. For each such term in the translation hypotheses, we compute the prevalence of one gender $g_{1}$ (e.g., ``\textit{diventata}''\textsuperscript{F}) over the other   $g_{2}$ (e.g., ``\textit{diventato}''\textsuperscript{M}) in the training corpus:
\begin{equation}
\textrm{Prevalence}(w_{g_1}, w_{g_2}) = \frac{\#w_{g_1}}{\#w_{g_1} + \#w_{g_2}}
\end{equation}
where $\#w$ denotes the number of occurrences of word form $w$ in the training data. 
We then compare the prevalence of term $w$ in gender $g_1$ with the model's preference for that gender when generating $w_{g_1}$. We quantify this preference by computing the relative probabilities between gendered alternatives:
\begin{equation}
\textrm{Preference}(w_{g_1}, w_{g_2}) = \frac{p(w_{g_1})}{p(w_{g_1}) + p(w_{g_2})}
\label{eq:pref}
\end{equation}
where $p(w)$ is the probability assigned by a given model to word $w$ in the translation hypothesis. 
This can be calculated as either the predicted gender preference (comparing the generated form against its ungenerated alternative) or the masculine preference (comparing masculine versus feminine forms regardless of which was generated).
By comparing prevalences with the model's gender preferences, we can distinguish predictions that replicate term-specific training patterns (where the higher-probability gender matches the more prevalent one in training data) from those based on acoustic or content-based source information, or other sources of bias.

\subsection{ILM Approximation}
\label{method:ilm}

While training data patterns provide one source of gender bias, the model's entrenched biases encompass more than only term-specific associations.
The decoder's behavior also reflects its learned understanding of target language structure, general patterns of gender marking, and how previously generated tokens constrain subsequent predictions. As the decoder autoregressively predicts tokens based on both encoder input and previously generated tokens, it develops these broader linguistic patterns, forming an internal language model.

To capture all aspects of these ingrained preferences that exist independently of the source audio, we analyze the ILM. Methods for its estimation have initially been developed for domain adaptation in ASR models \cite{variani2020hybrid,meng2021internal,zeineldeen2021investigating}.
We adopt the ILM estimation method of \citet{variani2020hybrid} and \citet{meng2021internal}, which replaces the encoder output with a dummy zero vector and has been shown to perform on par with more complex methods \cite{zeineldeen2021investigating}.
While \citet{fucci2023integrating} used the ILM to nudge ST models toward specific gender forms, here we analyze it to understand the decoder's inherent biases.

We compute the preference metric (Eq.~\ref{eq:pref}) for the ILM and compare it with the full model's preferences. Contrasting ILM and full model preferences reveals to what extent models follow these entrenched biases and when, conversely, the audio input overrides them.

\subsection{Feature Attribution}
\label{method:feat-attribution}

\begin{figure}
    \centering
    \includegraphics[width=\linewidth]{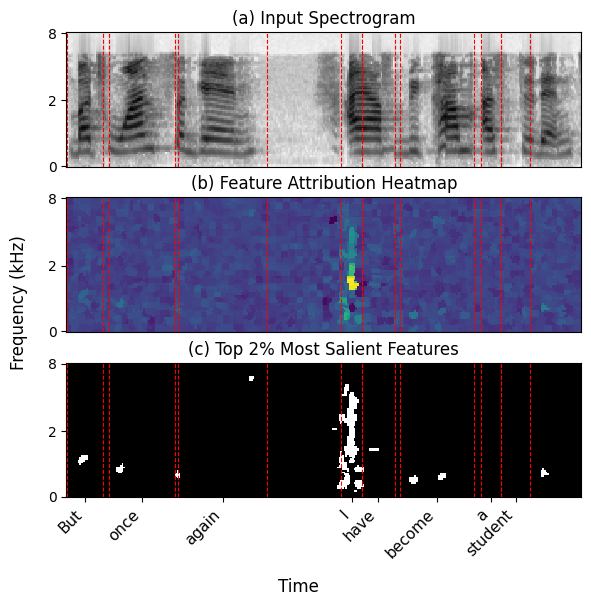}
    \caption{Example of contrastive feature attribution for the translation of ``\textit{become}'' to ``\textit{diventata}''$^F$ instead of ``\textit{diventato}''$^M$. (a) Input spectrogram. (b) Saliency heatmap showing features driving feminine gender assignment. (c) Top 2\,\% features sufficient to flip gender prediction.}
    \label{fig:example}
\end{figure}

Understanding how models use input audio requires dedicated interpretability methods. Existing approaches for speech-to-text models \cite{trinh2020directly, markertvisualizing, mohebbi2023homophone, wu2023explanations, fucci2024spes, wu2024can} primarily use perturbation techniques that mask input portions and measure the effect on model output.
However, these methods generate holistic explanations that highlight features relevant for all aspects of word generation. Since our goal is to identify which input features drive gender assignment specifically, we employ the contrastive feature attribution method of \newcite{conti2025unheard}.

This approach identifies why the model generates one gendered form instead of its alternative by computing relative probability changes between the two options when different parts of the input are masked.
Specifically, it automatically segments the input spectrogram (Figure~\ref{fig:example}.a) into acoustically meaningful regions and performs multiple inference passes with random perturbations to each segment. It assigns each segment a score based on how its perturbation affects the probability of one gendered form versus the other, producing saliency maps (Figure~\ref{fig:example}.b) that highlight the spectrogram regions most responsible for the model's gender choice. 

Following \citet{conti2025unheard}, we validate that the highlighted features are causally involved in gender assignment by testing whether occluding them changes the model's gender prediction. Figure~\ref{fig:example}.c shows the top 2\,\% of salient features that, when masked, successfully flip the prediction in this example. 
By occluding 1–20\,\% of the most salient features, we can flip the predicted gender in 37–47\,\% of examples across languages and models.
We focus our analysis on such flipped examples, where causal links between input features and gender assignment are established.
From these validated explanations, we can analyze which regions of the input spectrogram drive gender assignment: along the frequency dimension to identify relevant acoustic features, and along the time dimension to identify relevant words.

\section{Experimental Setup}
\label{sec:exp-setup}

\lc{We describe below the data, models, and evaluation setup used throughout our analyses.\footnote{Code to reproduce all analyses is available at \url{https://github.com/lina-conti/voice-bias-coreference}.}}

\paragraph{Data}

We use MuST-SHE \cite{bentivogli-etal-2020-gender}, a benchmark containing annotations for gender-neutral English terms in natural speech that require gender marking when translated to Spanish, French, or Italian.
We focus on the subset containing terms referring to the speaker, as these are cases where acoustic gender cues could influence gender assignment. Unlike sentences with gendered pronouns like ``\textit{She is a student},'' where gender is explicitly marked in the source, speaker-referential sentences like ``\textit{I am a student}'' contain no linguistic gender information, making acoustic cues potentially relevant.
For each term, the dataset provides correct and incorrect gender translations, e.g., ``\textit{diventata}''$^F$ vs. ``\textit{diventato}''$^M$ as Italian translations of ``\textit{become},'' which we use as contrastive pairs for our analysis. Following \citet{savoldi-etal-2022-morphosyntactic}, we exclude gender articles, as their high frequency in both genders across sentences makes it difficult to identify instances specifically referring to the speaker. We analyze only terms where the ST model generates one of the MuST-SHE annotated forms. Depending on the model used to translate, this yields between 309 and 453 gender terms per target language \lc{(see \S\ref{par:dataset-size} for details)}. 

\paragraph{Models}

We select models trained exclusively on a single open-source dataset to enable our training data analysis in \S\ref{sec:training-data}. We therefore focus on two model architectures trained on MuST-C \cite{cattoni2021must}: the multilingual Transformer encoder-decoder model by \citet{wang2020fairseq}, and the monolingual Conformer encoder-Transformer decoder models from \citet{papi2024good}. More recent ST systems and 
speech-enhanced large language models are typically trained on massive datasets that are not publicly released, making it difficult to establish connections between training data patterns and model behavior.

We primarily focus on the Transformer model for our analyses, as it demonstrates strong gender accuracy\footnote{The proportion of correct gender realizations among terms where the model generates one of the MuST-SHE annotated forms \cite{gaido-etal-2020-breeding}. \lc{Full results on gender accuracy for all models and language pairs are reported in Appendix~\ref{app:acc}.}}
for speaker-referential terms: 77.1\,\% to 80.6\,\% for feminine terms and 91.4\,\% to 94.4\,\% for masculine terms across target languages.
This suggests that the Transformer is well positioned to leverage vocal cues for gender disambiguation, a phenomenon we aim to investigate further\lc{.}
By comparing these results with Conformer models, which achieve lower accuracy (39.2-49.8\% for feminine terms and 72.5-76.7\% for masculine terms across the three language pairs), we assess whether gender assignment strategies are model-dependent.
These models provide architectural (Transformer vs. Conformer encoders) and scope (multilingual vs. monolingual) diversity, enabling us to examine how gender assignment strategies vary across different ST system configurations.

\section{Training Data Analysis}
\label{sec:training-data}

This section addresses our research question ``What is the influence of gender associations learned from the training data?'' Specifically, we measure whether models replicate term-specific gender patterns from their training data by computing gender prevalence (\S \ref{method:training-data}) in MuST-C \cite{cattoni2021must}, the training corpus for the models we analyze.

The first thing we observe is that the training data shows a clear masculine skew for the gender terms we study.
If we compute the average prevalence in the training data of the masculine form over the feminine one for all speaker-referring gendered terms in the translation hypotheses of the Transformer model, this average ranges from 0.68 to 0.71 depending on the languages \lc{(es: 0.68; fr: 0.71; it: 0.68)}, with nearly identical values for the Conformer models \lc{(es: 0.68; fr: 0,72; it: 0,68)}.

\begin{table}[!ht]
\small
\begin{center}
\begin{subtable}{0.32\textwidth}
\centering
\begin{tabular}{|c|c|c|}
\cline{2-3}
\multicolumn{1}{c|}{} & More Freq. & Less Freq. \\
\hline
F & 24 & 173 \\
M & 221 & 22 \\
\hline
\end{tabular}
\caption{Spanish}
\end{subtable}
\hfill
\begin{subtable}{0.32\textwidth}
\centering
\begin{tabular}{|c|c|c|}
\cline{2-3}
\multicolumn{1}{c|}{} & More Freq. & Less Freq. \\
\hline
F & 22 & 130 \\
M & 187 & 16 \\
\hline
\end{tabular}
\caption{French}
\end{subtable}
\hfill
\begin{subtable}{0.32\textwidth}
\centering
\begin{tabular}{|c|c|c|}
\cline{2-3}
\multicolumn{1}{c|}{} & More Freq. & Less Freq. \\
\hline
F & 12 & 140 \\
M & 192 & 13 \\
\hline
\end{tabular}
\caption{Italian}
\end{subtable}
\caption{Distribution of examples by predicted gender and whether the predicted gender is more or less prevalent in the training data for that term. Results for the Transformer model \cite{wang2020fairseq}.}
\label{tab:training-data}
\end{center}
\end{table}

Table~\ref{tab:training-data} shows that the Transformer model does not simply replicate training data patterns when assigning gender. 
The table categorizes each predicted gendered term by whether the predicted gender (F or M) is the more or less prevalent form in the training data for that specific term. If models followed the heuristic of generating each term in its most frequent training data gender, predictions should consistently fall in the “More Freq.” column.
Instead, the model frequently predicts genders that are less prevalent for that specific term: 87.8\,\% of feminine generated terms in Spanish (173 of 197), 85.5\,\% in French (130 of 152), and 92.1\,\% in Italian (140 of~152) correspond to the less prevalent form in the training data. For masculine predictions, the model does tend to predict the more prevalent form (221 of 243 in Spanish, 187 of 203 in French, 192 of 205 in Italian). However, given the overall masculine skew, this reflects the general pattern in the training data rather than term-specific memorization. The Conformers show a similar pattern \lc{(see Table~\ref{tab:training-data-conf})}: the distribution of predictions relative to prevalence resembles Table~\ref{tab:training-data}, although with slightly more predictions aligning with the more prevalent form.
Crucially, none of the models closely follow term-by-term gender associations from the training data.

In summary, our results challenge the assumption that gender bias in ST simply reflects the training data distribution \cite{tatman2017gender, garnerin2019gender, iluz-etal-2023-exploring, mastromichalakis2025assumed}. 
Our findings align with \citet{conti2023using} and \citet{elghazaly2025exploring}, suggesting that gender bias cannot be exclusively reduced to training data imbalances. 
The data's masculine skew clearly influences model behavior, but not through simple memorization—rather, models internalize broader biases that we investigate through the ILM.

\section{Internal Language Model Analysis}
\label{sec:ilm}

While the training data analysis has revealed that models do not simply memorize term-specific associations, gender assignment must still be driven by some combination of learned decoder preferences and input audio features. 
We first investigate whether the decoder has internalized broader biases beyond individual term associations by addressing our second research question: ``What is the impact of the model's learned knowledge of the target language and a priori assumptions about gender on predictions?''
The ILM analysis isolates these entrenched preferences by removing encoder information, measuring what the decoder learned\lc{,} independently of source audio. Comparing ILM \lc{predictions} with full model predictions then reveals when and how the audio input overrides these biases.

The ILM reflects and amplifies the masculine skew observed in the training data. Averaging over all gender terms, the ILM's preference for masculine over feminine ranges from 0.74 to 0.81 for the Transformer, depending on \lc{the} language---higher than the training data prevalence of 0.68--0.71. 
When we separate by the gender that is ultimately generated by the full model, the ILM's masculine preference is 0.85--0.88 for masculine predictions and 0.58--0.71 for feminine ones (always above 0.5, even when generating feminine forms; \lc{see Appendix~\ref{app:ilm} for full results}).
For the Conformer models, average masculine preference is 0.63--0.64 (0.71--0.74 for masculine predictions, 0.48--0.49 for feminine ones). While the Conformers' masculine preference drops just below 0.5 for feminine predictions, for masculine ones it remains well above 0.5, suggesting some masculine bias, though weaker than for the Transformer.

\begin{table}[!ht]
\small
\begin{center}
\begin{subtable}{0.32\textwidth}
\centering
\begin{tabular}{|c|c|c|}
\cline{2-3}
\multicolumn{1}{c|}{} & Higher Prob. & Lower Prob. \\
\hline
F & 52 & 145 \\
M & 225 & 18 \\
\hline
\end{tabular}
\caption{Spanish}
\end{subtable}
\hfill
\begin{subtable}{0.32\textwidth}
\centering
\begin{tabular}{|c|c|c|}
\cline{2-3}
\multicolumn{1}{c|}{} & Higher Prob. & Lower Prob. \\
\hline
F & 33 & 119 \\
M & 195 & 8 \\
\hline
\end{tabular}
\caption{French}
\end{subtable}
\hfill
\begin{subtable}{0.32\textwidth}
\centering
\begin{tabular}{|c|c|c|}
\cline{2-3}
\multicolumn{1}{c|}{} & Higher Prob. & Lower Prob. \\
\hline
F & 46 & 106 \\
M & 195 & 10 \\
\hline
\end{tabular}
\caption{Italian}
\end{subtable}
\caption{Distribution of examples by predicted gender and whether the ILM assigns higher or lower probability to the predicted gender compared to the alternative. Results for the Transformer model \cite{wang2020fairseq}.}
\label{tab:ilm}
\end{center}
\end{table}

However, the full model frequently overrides these entrenched biases. In the example in Figure~\ref{fig:example}, the prevalence in training data for masculine ``\textit{diventato}'' is 0.57, and the ILM preference is 0.85, yet the full model's preference for the predicted feminine form ``\textit{diventata}'' is 0.99, illustrating how acoustic input can supersede learned biases and training data statistics. 
Table~\ref{tab:ilm} shows this is common: the Transformer frequently predicts genders to which the ILM assigns lower probability, particularly for feminine predictions. Instead, the Conformers seem to rely more on their ILM: the Pearson correlation between ILM and full model masculine preference is strong for Conformers (es: 0.65; fr: 0.62; it: 0.62), but weak to moderate for the Transformer (es: 0.45; fr: 0.38; it: 0.47).

This section has shown that the ILM exhibits a masculine-as-norm bias across our models, but ST systems vary in how much they rely on these entrenched preferences versus input audio. The Transformer's ILM is strongly biased toward masculine, yet the full model frequently overrides these preferences based on acoustic information. The Conformers show weaker ILM masculine bias but rely more on it, struggling to leverage source audio effectively. This analysis demonstrates that ST models combine acoustic gender cues with language model preferences to varying degrees. 
These findings demonstrate that acoustic input can play a substantial role in gender assignment, motivating us to investigate which specific aspects of the audio our models exploit for this.

\section{The Role of Input Audio}
\label{sec:feat-attribution}

This section addresses our third research question: ``What aspects of the input audio does the model use to assign gender to terms referring to the speaker?'' 
For this, we apply the feature attribution method from \S \ref{method:feat-attribution}. 
Occluding 1--20\,\% of the most salient features highlighted by the saliency map flips the predicted gender in 40.7\,\% of Spanish examples, 46.8\,\% of French examples, and 37.0\,\% of Italian examples for the Transformer model, with comparable rates for the Conformers \lc{(es: 41.1\,\%; fr: 44.9\,\%; it: 43.0\,\%)}.
For these flipped examples, we have a guarantee that the highlighted features are causally involved in gender assignment, since if we occlude them, the model's prediction changes.
We analyze these saliency maps to identify which words and acoustic cues in the source audio influence gender assignment.

\subsection{Frequency Dimension}

\begin{figure*}
    \centering
    \includegraphics[width=1\linewidth]{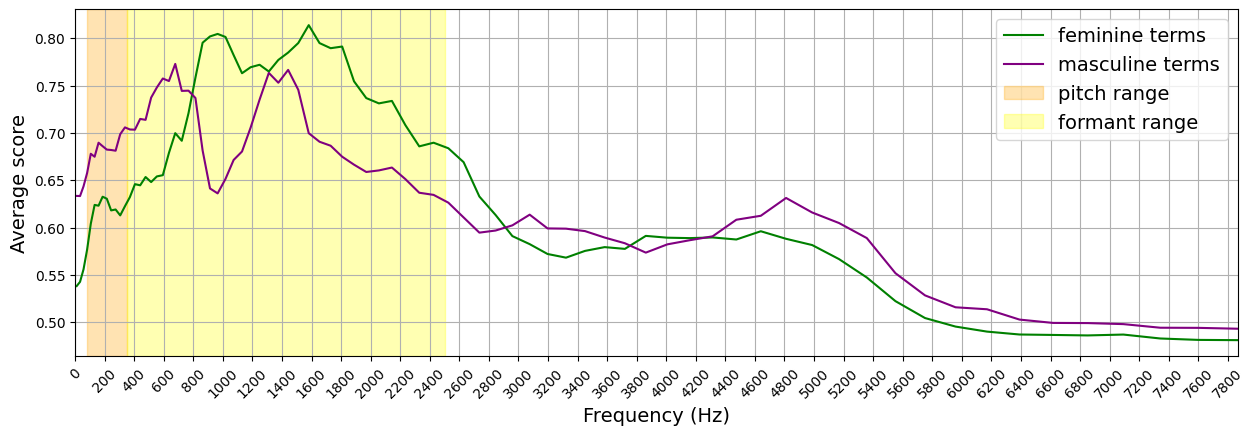}
    \caption{
    \lc{Average saliency scores per frequency bin (max-pooled over the time dimension, then averaged across all gender terms that flip), for the Transformer model \cite{wang2020fairseq} on en→it translation. Results are shown separately for feminine and masculine terms. Shaded regions mark the pitch range (80--350~Hz) and formant range (350--2500~Hz).}}
    \label{fig:frequency-scores}
\end{figure*}

Prior work assumes that, since pitch is strongly associated with perceived gender, 
ST models use it when correctly assigning gender to terms referring to the speaker \cite{bentivogli-etal-2020-gender, elaraby2018gender, fucci2023no}. We empirically test this assumption for the first time.

Pitch is the perceptual correlate of the fundamental frequency F$_0$ \cite{dan2009speech}: utterances with higher F$_0$ sound higher pitched and more feminine, whereas male speech typically has lower F$_0$ \cite{simpson2009phonetic}. To determine whether the model relies on pitch, we examine the intensity with which the pitch region (80-350\,Hz, where the fundamental frequency lies) is highlighted in our heatmaps. We aggregate each gender term's explanation by taking the max score for each frequency bin over the time dimension, then average across all gender terms. 

Surprisingly, the pitch range does not show the highest scores, suggesting it is not the most important region driving the choice of gender to refer to the speaker for the models we study. Figure~\ref{fig:frequency-scores} shows the average score across the frequency range for the Transformer model on the en$\rightarrow$it split, with the same pattern holding for other languages and for the Conformer models \lc{(see Figure~\ref{fig:frequency-scores-full})}.
Instead, the formant range (350--2,500\,Hz) displays the highest scores, with two peaks corresponding to the first and second formants (F$_1$ and F$_2$). These formants are important for identifying the word being uttered, especially vowels \cite{dan2009speech}, but their exact frequency also varies by speaker and, notably, depending on the speaker's gender \cite{simpson2009phonetic}. This is visible in Figure~\ref{fig:frequency-scores}:
\lc{for masculine terms, saliency peaks appear at approximately 500--900\,Hz and 1,200--1,600\,Hz, while for feminine terms the peaks shift to approximately 800--1,100\,Hz and 1,400--1,800\,Hz, broadly consistent with the ranges of F$_1$ and F$_2$ observed in English vowels for male and female speakers respectively \cite{hillenbrand2009role}. }
This suggests that the assumption that ST models should leverage pitch information to disambiguate the gender of terms referring to the speaker does not fully correspond to our model's behavior.

Still, while not the dominant feature, we cannot exclude that pitch plays a role in the model's decision process.
In the example in Figure~\ref{fig:example}, occluding the top 2\,\% of features with the highest scores flips the gender of the translation of ``\textit{become}'' from feminine to masculine. These features are concentrated along the time axis but spread across most of the frequency range and, crucially, they include the pitch region. This pattern holds for all our samples: 99.9\,\% of examples that flip contain at least one feature in the pitch 
\lc{range} among those occluded for flipping (99.8\,\% for the Conformers).

Overall, we can conclude that the information \lc{that} the model uses is distributed across the frequency range rather than concentrated in pitch alone, with particular emphasis on F$_1$ and F$_2$. This has implications for interventions to mitigate gender bias or neutralize the gender of audio samples, suggesting that approaches acting solely on pitch would be insufficient.
Moreover, since salient features for gender disambiguation are concentrated along the time dimension but distributed along the frequency dimension, this suggests that \textit{when} acoustic cues occur may be particularly important. We now turn to analyzing which source words correspond to these temporally concentrated features. 

\subsection{Time Dimension}

\begin{table}[!ht]
\small

    \centering
    \begin{tabular}{c|ccc|ccc|}
        \cline{2-7}
        & \multicolumn{3}{c|}{`\textit{I}'} & \multicolumn{3}{c|}{Self-referential} \\
        \cline{2-7}
         & es & fr & it & es & fr & it \\
         \hline
       \multicolumn{1}{|c|}{Flip} & \lc{38.0} & \lc{36.1} & \lc{46.2} & \lc{45.2} & \lc{43.4} & \lc{53.8}\\
       \multicolumn{1}{|c|}{All} & 25.4 & 28.4 & 33.3 & 32.5 & 35.2 & 42.0 \\
        \hline
    \end{tabular}
    \caption{Percentage of examples where the top-scoring source word is ``\textit{I}'' or a self-referential word, for examples that flip and for all examples. Results for the Transformer model \cite{wang2020fairseq}.}
    \label{tab:self-ref-percent}
\end{table}

Analyzing which source words drive gender assignment reveals a surprising pattern: models rely primarily on first-person pronouns and determiners that refer to the speaker.
Manual investigation of the contrastive heatmaps has revealed that ``\textit{I}'' is the word most frequently highlighted to explain gender choice. To validate this observation quantitatively, we extract word-level scores from the spectrogram heatmaps by using Gentle\footnote{https://github.com/strob/gentle} to obtain each source word's time range and selecting the highest feature score within that range as the word's score. 
\lc{Table~\ref{tab:self-ref-percent} shows that, for the Transformer model, ``\textit{I}'' is the top-scoring word in 36.1--46.2\,\% of examples that flip, depending on the language. Including other self-referential expressions (``\textit{I}'', ``\textit{I'd}'', ``\textit{I've}'', ``\textit{I'm}'', ``\textit{my}'', ``\textit{me}'', ``\textit{myself}'') raises these percentages to 43.4--53.8\,\%. Crucially, these percentages are higher for examples that flip than for all examples (second row of Table~\ref{tab:self-ref-percent}), suggesting that self-referential words are not merely frequent top-scoring features, but are causally involved in gender assignment: occluding them can be sufficient to change the model's prediction.}
Besides these first-person words, Table~\ref{tab:top-words} shows that words like 
\lc{``\textit{was}'', ``\textit{when}'', and ``\textit{and}'}
also frequently receive high attribution scores. However, manual inspection reveals these words appear next to ``\textit{I}'' in the source sentence (``\textit{and I...}'', ``\textit{I was...}'', ``\textit{when I...}'') and likely score highest due to imprecisions in Gentle's word-level alignments, suggesting the actual prevalence of first-person words is even greater.

\begin{table}[!ht]
\small

\begin{center}
\begin{tabularx}{\columnwidth}{|lX|lX|lX|}
\hline
\multicolumn{2}{|c|}{es} & \multicolumn{2}{c}{fr} & \multicolumn{2}{|c|}{it} \\
\hline
I & 40 & I & 47 & I & 45 \\
I'm & 20 & was & 7 & I'm & 9 \\
as & 15 & I'm & 6 & I've & 5 \\
was & 6 & when & 6 & me & 4 \\
and & 4 & and & 6 & and & 3 \\
engineer & 3 & I've & 5 & was & 3 \\
myself & 3 & know & 3 & grew & 3 \\
that & 3 & wasn't & 2 & as & 2 \\
my & 3 & as & 2 & that & 2 \\
when & 3 & so & 2 & know & 2 \\
\hline
\end{tabularx}
\caption{Most frequent top-scoring source words for examples that flip, with the number of examples \lc{for which they receive the highest saliency score in the source sentence.} Results for the Transformer model \cite{wang2020fairseq}.}
\label{tab:top-words}
\end{center}
\end{table}

The reliance on first-person pronouns differs across model architectures. 
For the Conformers, self-referential words are top-scoring in only 
\lc{14.7--19.5\,\%}
of flipped examples \lc{(see Table~\ref{tab:self-ref-percent-conf})}, compared to 
\lc{43.4--53.8\,\%}
for the Transformer. Despite this quantitative difference, 
\lc{the pronoun ``\textit{I}''}
consistently appears at the top across all models and languages \lc{(Table~\ref{tab:top-words-conformer})}, indicating that this strategy is learnable by different architectures but exploited with varying effectiveness. The Transformer, which achieves higher gender translation accuracy, relies on this mechanism more frequently than the Conformers. This pattern 
\lc{concurs with the finding of \citet{fucci-etal-2025-different}} that models with higher gender translation accuracy for speaker-referring terms also encode more gender information in their representations (measured via probing). Our contrastive analysis goes further by revealing the mechanism through which models access this information: via self-referential words that establish coreference with the speaker.

Beyond architectural differences, models' reliance on input features also differs between feminine and masculine predictions. For the Transformer, the percentage of examples whose gender prediction we can flip by occluding salient features in the input spectrogram is 60.8\,\% vs. 19.2\,\% for Spanish, 66.7\,\% vs. 24.9\,\% for French, and 52.2\,\% vs. 21.1\,\% for Italian (feminine vs. masculine respectively). The Conformers show the same pattern with a narrower gap (es: 47.5\,\% vs. 33.6\,\%; fr: 51.0\,\% vs. 37.6\,\%; it: 44.7\,\% vs. 41.3\,\%). This asymmetry suggests that the mechanism of accessing acoustic gender cues through first-person pronouns plays a more critical role for feminine predictions, while masculine predictions rely more heavily on the model's internal biases. This asymmetry aligns with prior work showing that language models and MT systems use masculine as a default, requiring strong feminine signals to generate feminine forms \cite{jumelet2019analysing, manna-etal-2025-paying}, which they slowly and imperfectly learn to use during training \cite{savoldi2022dynamics}.

This reliance on first-person pronouns reflects a mechanism analogous to coreference resolution in text-based MT.
Just as MT models rely on gendered pronouns and determiners (e.g., ``\textit{she}''/``\textit{he}'', ``\textit{her}''/``\textit{his}'') that corefer with gender-ambiguous terms for disambiguation \cite{voita-etal-2018-context, escude-font-costa-jussa-2019-equalizing, stanovsky-etal-2019-evaluating, manna-etal-2025-paying}, our ST models rely on first-person words that refer to the speaker. 
However, while gendered pronouns in text carry explicit gender information, words like ``\textit{I}'' are semantically gender-neutral. 
In ST, acoustic gender cues in the speaker's voice effectively transform these semantically neutral words into 
gendered markers. Through coreference with the speaker, ``\textit{I}'' provides access to the same gender information that explicit gendered pronouns provide in text-based MT, but encoded acoustically rather than lexically.

Besides first-person pronouns, models sometimes assign high salience to the source words corresponding to the gendered target terms. 
Table~\ref{tab:top-words} contains examples of this phenomenon: words like 
\lc{``\textit{grew}'', which translates to ``\textit{cresciuto}''$^M$ or ``\textit{cresciuta}''$^F$} 
in Italian. These words are directly relevant for translating the gendered term itself and, like all words in the utterance, carry acoustic gender cues that models can access.

In summary, the contrastive saliency maps reveal that ST models frequently rely on self-referential words like ``\textit{I}'' to establish coreference with the speaker, enabling access to acoustic gender cues.
These cues are distributed across the frequency spectrum rather than concentrated in pitch, with $F_1$ and $F_2$ showing higher importance.

\section{Conclusion}

We investigated how ST models assign gender to speaker-referring terms when translating from English to three Romance languages. 
Our analysis revealed that rather than memorizing individual term-gender pairings from training data, models learn that masculine forms are generally more prevalent. The decoder exhibits strong bias toward masculine defaults independent of input audio, but information from the audio can override these preferences. Crucially, models leverage first-person pronouns analogously to gendered pronouns in MT: acoustic cues transform the semantically neutral "I" into a functionally gendered marker, encoding gender information primarily in formant frequencies rather than pitch.
These findings demonstrate that ST models can use vocal cues for gender disambiguation through mechanisms distinct from those assumed in prior work. They suggest that mitigation strategies centered on
pitch manipulation \cite{fucci2023no} or exclusively rebalancing training data \cite{garnerin2019gender, bansal2025addressing} may prove insufficient, as they fail to account for the distributed nature of acoustic gender cues and how models mediate training statistics through complex learning dynamics.

\section{Ethics Statement}
\label{sec:ethics-statement}

\paragraph{Broader Impact.} To mitigate harmful behaviors in AI systems such as those outlined in our Bias Statement (\S\ref{sec:bias-statement}), we need both mitigation strategies \cite{vanmassenhove2018getting, escude-font-costa-jussa-2019-equalizing, Saunders2020ReducingGBA, saunders2022first} and foundational research that reveals the mechanisms underlying biased behaviors. This interpretability work contributes to the latter effort, providing insights into how speech translation models make gender assignments that can inform future interventions.

\paragraph{Binary Gender Framework.} Our analytical framework requires ethical consideration. We work within a binary gender framework, which offers methodological advantages---enabling contrastive explanations and leveraging existing benchmarks---but comes with significant limitations. This binary approach fails to account for gender identities outside the male/female binary \cite{zimman-2020-transgender-lang}, potentially contributing to their erasure \cite{calado-2025-myths}. Our contrastive analysis compares feminine versus masculine term generation without considering other possibilities such as gender-neutral reformulations \cite{piergentili2023hi, savoldi2025mgente} or neologisms that avoid the binary dichotomy \cite{piergentili2024neogate}. While existing ST benchmarks like MuST-SHE \cite{bentivogli-etal-2020-gender} provide binary gender annotations, comparable annotations for non-binary alternatives in speech do not yet exist. With such resources, we could potentially extend our analysis to include these forms, even though models rarely generate them spontaneously. We acknowledge that this resource-driven limitation means our analysis cannot capture the full spectrum of gender identities.

\paragraph{Gender Inference from Vocal Cues.} Our analysis examines whether models use vocal cues for gender assignment, which could improve accuracy on binary benchmarks like MuST-SHE \citep{bentivogli-etal-2020-gender}. However, we do not advocate that models should rely on vocal characteristics to infer gender, as this risks equating gender with sex and treating it as biologically determined rather than as the behavioral and social phenomenon it is \citep{butler1990feminism}. Such an approach could harm speakers whose voices do not align with binary gender expectations. Instead, our goal is to understand what information current models exploit and how. Understanding these mechanisms provides a foundation for developing better control over model behavior, enabling different approaches depending on context and user preferences---whether that involves automatic inference, gender-neutral translations in ambiguous cases, or respecting user-specified pronouns and gender identity.

\section{Limitations}

\paragraph{Models.} Our analysis focuses on two model architectures trained on the MuST-C dataset \cite{cattoni2021must}. While these models are not state-of-the-art in terms of overall speech translation performance, we selected them for specific methodological reasons. The Transformer model \cite{wang2020fairseq} demonstrates notably higher accuracy on gender assignment compared to more recent systems like SeamlessM4T \cite{seamless2025joint}, making it better positioned to reveal how models successfully leverage acoustic cues for gender disambiguation. Additionally, both models are trained exclusively on MuST-C, which enables the training data analysis in \S\ref{sec:training-data}. In contrast, modern speech translation systems and speech-enhanced large language models are typically trained on multiple large-scale datasets without transparent documentation, making it difficult to establish connections between training data patterns and model behavior. The models we analyze provide architectural diversity (Transformer versus Conformer encoders) and scope differences (multilingual versus monolingual), allowing us to examine how gender assignment strategies vary across different system configurations. However, they are trained with the same objective (supervised training with cross entropy) and we do not cover different training regimes such as non-autoregressive or self-supervised trainings. Additionally, our analysis does not extend to speech-enhanced large language models (SpeechLLMs) \cite{rubenstein2023audiopalm, wu2023decoder, tang2024salmonn}, which represent an emerging paradigm for speech translation with potentially different mechanisms for processing acoustic information and assigning gender. Future work should investigate whether the patterns we identified---particularly the use of first-person pronouns for coreference resolution, the distribution of salient features across the frequency spectrum rather than concentration in pitch, and the relationship between internal language model biases and full model predictions---generalize to recent models trained on large-scale data, to models trained with different objectives, and to SpeechLLMs.

\paragraph{Language Coverage.} 
Our analysis uses the MuST-SHE benchmark \cite{bentivogli-etal-2020-gender}, which covers the translation from English into three Romance languages (Spanish, French, and Italian). Since gender cues may manifest differently in other languages, this may influence how models learn gender assignment patterns. For this reason, future work should extend our analysis to typologically diverse language pairs, including languages with different gender marking strategies or non-gendered source languages. This would help determine whether the mechanisms we identify represent general model strategies or language-specific patterns.

\paragraph{Dataset Size.}
\label{par:dataset-size}
The benchmark also constrains our dataset size. After filtering as described in \S\ref{sec:exp-setup}, we obtain 440 gender terms for Spanish, 355 for French, and 357 for Italian for the Transformer model, and 453 gender terms for Spanish, 379 for French, and 309 for Italian for the Conformer models. While this represents a relatively small dataset, MuST-SHE is currently the only available resource for speech translation that provides the fine-grained annotations required for our interpretability methods: gender terms referring to the speaker that are gender-neutral in English but gendered in the target language, gold-standard gender labels reflecting speakers' self-identified gender, and contrastive gender alternatives for each term. These annotations are essential for the contrastive feature attribution method we employ \cite{conti2025unheard}. The consistency of findings across three language pairs strengthens confidence in our results despite the dataset size.

\section{Acknowledgments}

This paper has received funding from the PNRR project FAIR - Future AI Research (PE00000013), under the NRRP MUR program funded by the NextGenerationEU and from the European Union’s Horizon research and innovation programme under
grant agreement No 101135798, project Meetween (My Personal AI Mediator for Virtual MEETings BetWEEN People).

\section{Bibliographical References}\label{sec:reference}

\bibliographystyle{lrec2026-natbib}
\bibliography{mybib}

@article{fucci2024spes,
  title={SPES: Spectrogram perturbation for explainable speech-to-text generation},
  author={Fucci, Dennis and Gaido, Marco and Savoldi, Beatrice and Negri, Matteo and Cettolo, Mauro and Bentivogli, Luisa},
  journal={arXiv preprint arXiv:2411.01710},
  year={2024}
}

@inproceedings{fucci-etal-2025-different,
    title = "Different Speech Translation Models Encode and Translate Speaker Gender Differently",
    author = "Fucci, Dennis  and
      Gaido, Marco  and
      Negri, Matteo  and
      Bentivogli, Luisa  and
      Martins, Andre  and
      Attanasio, Giuseppe",
    editor = "Che, Wanxiang  and
      Nabende, Joyce  and
      Shutova, Ekaterina  and
      Pilehvar, Mohammad Taher",
    booktitle = "Proceedings of the 63rd Annual Meeting of the Association for Computational Linguistics (Volume 2: Short Papers)",
    month = jul,
    year = "2025",
    address = "Vienna, Austria",
    publisher = "Association for Computational Linguistics",
    url = "https://aclanthology.org/2025.acl-short.78/",
    doi = "10.18653/v1/2025.acl-short.78",
    pages = "1005--1019",
    ISBN = "979-8-89176-252-7"
}

@inproceedings{attanasio-etal-2023-tale,
    title = "A Tale of Pronouns: Interpretability Informs Gender Bias Mitigation for Fairer Instruction-Tuned Machine Translation",
    author = "Attanasio, Giuseppe  and
      Plaza del Arco, Flor Miriam  and
      Nozza, Debora  and
      Lauscher, Anne",
    editor = "Bouamor, Houda  and
      Pino, Juan  and
      Bali, Kalika",
    booktitle = "Proceedings of the 2023 Conference on Empirical Methods in Natural Language Processing",
    month = dec,
    year = "2023",
    address = "Singapore",
    publisher = "Association for Computational Linguistics",
    url = "https://aclanthology.org/2023.emnlp-main.243/",
    doi = "10.18653/v1/2023.emnlp-main.243",
    pages = "3996--4014"
}

@inproceedings{savoldi-etal-2022-morphosyntactic,
    title = "Under the Morphosyntactic Lens: A Multifaceted Evaluation of Gender Bias in Speech Translation",
    author = "Savoldi, Beatrice  and
      Gaido, Marco  and
      Bentivogli, Luisa  and
      Negri, Matteo  and
      Turchi, Marco",
    editor = "Muresan, Smaranda  and
      Nakov, Preslav  and
      Villavicencio, Aline",
    booktitle = "Proceedings of the 60th Annual Meeting of the Association for Computational Linguistics (Volume 1: Long Papers)",
    month = may,
    year = "2022",
    address = "Dublin, Ireland",
    publisher = "Association for Computational Linguistics",
    url = "https://aclanthology.org/2022.acl-long.127/",
    doi = "10.18653/v1/2022.acl-long.127",
    pages = "1807--1824"
}

@inproceedings{costa-jussa-etal-2022-evaluating,
    title = "Evaluating Gender Bias in Speech Translation",
    author = "Costa-juss{\`a}, Marta R.  and
      Basta, Christine  and
      G{\'a}llego, Gerard I.",
    editor = "Calzolari, Nicoletta  and
      B{\'e}chet, Fr{\'e}d{\'e}ric  and
      Blache, Philippe  and
      Choukri, Khalid  and
      Cieri, Christopher  and
      Declerck, Thierry  and
      Goggi, Sara  and
      Isahara, Hitoshi  and
      Maegaard, Bente  and
      Mariani, Joseph  and
      Mazo, H{\'e}l{\`e}ne  and
      Odijk, Jan  and
      Piperidis, Stelios",
    booktitle = "Proceedings of the Thirteenth Language Resources and Evaluation Conference",
    month = jun,
    year = "2022",
    address = "Marseille, France",
    publisher = "European Language Resources Association",
    url = "https://aclanthology.org/2022.lrec-1.230/",
    pages = "2141--2147"
}

@inproceedings{attanasio-etal-2024-twists,
    title = "Twists, Humps, and Pebbles: Multilingual Speech Recognition Models Exhibit Gender Performance Gaps",
    author = "Attanasio, Giuseppe  and
      Savoldi, Beatrice  and
      Fucci, Dennis  and
      Hovy, Dirk",
    editor = "Al-Onaizan, Yaser  and
      Bansal, Mohit  and
      Chen, Yun-Nung",
    booktitle = "Proceedings of the 2024 Conference on Empirical Methods in Natural Language Processing",
    month = nov,
    year = "2024",
    address = "Miami, Florida, USA",
    publisher = "Association for Computational Linguistics",
    url = "https://aclanthology.org/2024.emnlp-main.1188/",
    doi = "10.18653/v1/2024.emnlp-main.1188",
    pages = "21318--21340"
}

@inproceedings{fucci2023no,
  title={No pitch left behind: Addressing gender unbalance in automatic speech recognition through pitch manipulation},
  author={Fucci, Dennis and Gaido, Marco and Negri, Matteo and Cettolo, Mauro and Bentivogli, Luisa},
  booktitle={2023 IEEE Automatic Speech Recognition and Understanding Workshop (ASRU)},
  pages={1--8},
  year={2023},
  organization={IEEE}
}

@inproceedings{elaraby2018gender,
  title={Gender aware spoken language translation applied to english-arabic},
  author={Elaraby, Mostafa and Tawfik, Ahmed Y and Khaled, Mahmoud and Hassan, Hany and Osama, Aly},
  booktitle={2018 2nd International Conference on Natural Language and Speech Processing (ICNLSP)},
  pages={1--6},
  year={2018},
  organization={IEEE}
}

@inproceedings{jumelet2019analysing,
  title={Analysing Neural Language Models: Contextual Decomposition Reveals Default Reasoning in Number and Gender Assignment},
  author={Jumelet, Jaap and Zuidema, Willem and Hupkes, Dieuwke},
  booktitle={Proceedings of the 23rd Conference on Computational Natural Language Learning (CoNLL)},
  pages={1--11},
  year={2019}
}

@inproceedings{garnerin2019gender,
  title={Gender representation in French broadcast corpora and its impact on ASR performance},
  author={Garnerin, Mahault and Rossato, Solange and Besacier, Laurent},
  booktitle={Proceedings of the 1st international workshop on AI for smart TV content production, access and delivery},
  pages={3--9},
  year={2019}
}

@inproceedings{voita-etal-2018-context,
    title = "Context-Aware Neural Machine Translation Learns Anaphora Resolution",
    author = "Voita, Elena  and
      Serdyukov, Pavel  and
      Sennrich, Rico  and
      Titov, Ivan",
    editor = "Gurevych, Iryna  and
      Miyao, Yusuke",
    booktitle = "Proceedings of the 56th Annual Meeting of the Association for Computational Linguistics (Volume 1: Long Papers)",
    month = jul,
    year = "2018",
    address = "Melbourne, Australia",
    publisher = "Association for Computational Linguistics",
    url = "https://aclanthology.org/P18-1117/",
    doi = "10.18653/v1/P18-1117",
    pages = "1264--1274"
}

@inproceedings{iluz-etal-2023-exploring,
    title = "Exploring the Impact of Training Data Distribution and Subword Tokenization on Gender Bias in Machine Translation",
    author = "Iluz, Bar  and
      Limisiewicz, Tomasz  and
      Stanovsky, Gabriel  and
      Mare{\v{c}}ek, David",
    editor = "Park, Jong C.  and
      Arase, Yuki  and
      Hu, Baotian  and
      Lu, Wei  and
      Wijaya, Derry  and
      Purwarianti, Ayu  and
      Krisnadhi, Adila Alfa",
    booktitle = "Proceedings of the 13th International Joint Conference on Natural Language Processing and the 3rd Conference of the Asia-Pacific Chapter of the Association for Computational Linguistics (Volume 1: Long Papers)",
    month = nov,
    year = "2023",
    address = "Nusa Dua, Bali",
    publisher = "Association for Computational Linguistics",
    url = "https://aclanthology.org/2023.ijcnlp-main.57/",
    doi = "10.18653/v1/2023.ijcnlp-main.57",
    pages = "885--896"
}

@Book{dan2009speech,
  title={Speech and language processing: An introduction to natural language processing, computational linguistics, and speech recognition},
  author={Jurafsky, Dan and James, Martin H.},
  year={2009},
  publisher={Pearson Prentice Hall Upper Saddle River, NJ}
}

@inproceedings{blodgett-etal-2020-language,
    title = "Language (Technology) is Power: A Critical Survey of ``Bias'' in {NLP}",
    author = "Blodgett, Su Lin  and
      Barocas, Solon  and
      Daum{\'e} III, Hal  and
      Wallach, Hanna",
    editor = "Jurafsky, Dan  and
      Chai, Joyce  and
      Schluter, Natalie  and
      Tetreault, Joel",
    booktitle = "Proceedings of the 58th Annual Meeting of the Association for Computational Linguistics",
    month = jul,
    year = "2020",
    address = "Online",
    publisher = "Association for Computational Linguistics",
    url = "https://aclanthology.org/2020.acl-main.485/",
    doi = "10.18653/v1/2020.acl-main.485",
    pages = "5454--5476"
}

@article{cattoni2021must,
  title={Must-c: A multilingual corpus for end-to-end speech translation},
  author={Cattoni, Roldano and Di Gangi, Mattia Antonino and Bentivogli, Luisa and Negri, Matteo and Turchi, Marco},
  journal={Computer speech \& language},
  volume={66},
  pages={101155},
  year={2021},
  publisher={Elsevier}
}

@inproceedings{bentivogli-etal-2020-gender,
    title = "Gender in Danger? Evaluating Speech Translation Technology on the {M}u{ST}-{SHE} Corpus",
    author = "Bentivogli, Luisa  and
      Savoldi, Beatrice  and
      Negri, Matteo  and
      Di Gangi, Mattia A.  and
      Cattoni, Roldano  and
      Turchi, Marco",
    editor = "Jurafsky, Dan  and
      Chai, Joyce  and
      Schluter, Natalie  and
      Tetreault, Joel",
    booktitle = "Proceedings of the 58th Annual Meeting of the Association for Computational Linguistics",
    month = jul,
    year = "2020",
    address = "Online",
    publisher = "Association for Computational Linguistics",
    url = "https://aclanthology.org/2020.acl-main.619/",
    doi = "10.18653/v1/2020.acl-main.619",
    pages = "6923--6933"
}

@article{
doi:10.1073/pnas.1915768117,
author = {Allison Koenecke  and Andrew Nam  and Emily Lake  and Joe Nudell  and Minnie Quartey  and Zion Mengesha  and Connor Toups  and John R. Rickford  and Dan Jurafsky  and Sharad Goel },
title = {Racial disparities in automated speech recognition},
journal = {Proceedings of the National Academy of Sciences},
volume = {117},
number = {14},
pages = {7684-7689},
year = {2020},
doi = {10.1073/pnas.1915768117},
URL = {https://www.pnas.org/doi/abs/10.1073/pnas.1915768117},
eprint = {https://www.pnas.org/doi/pdf/10.1073/pnas.1915768117}}

@article{chao2021girl,
  title={Girl talk: Understanding negative reactions to female vocal fry},
  author={Chao, Monika and Bursten, Julia RS},
  journal={Hypatia},
  volume={36},
  number={1},
  pages={42--59},
  year={2021},
  publisher={Cambridge University Press}
}

@article{brown2025sociophonetic,
  title={A sociophonetic study of creaky voice across language, gender and age in Canadian English-French bilinguals},
  author={Brown, Jeanne and Sonderegger, Morgan},
  journal={Journal of Phonetics},
  volume={112},
  pages={101431},
  year={2025},
  publisher={Elsevier}
}

@article{elghazaly2025exploring,
  title={Exploring Gender Disparities in Automatic Speech Recognition Technology},
  author={Elghazaly, Hend and Mirheidari, Bahman and Moosavi, Nafise Sadat and Christensen, Heidi},
  journal={CoRR},
  year={2025}
}

@inproceedings{tatman2017gender,
  title={Gender and dialect bias in YouTube’s automatic captions},
  author={Tatman, Rachael},
  booktitle={Proceedings of the first ACL workshop on ethics in natural language processing},
  pages={53--59},
  year={2017}
}

@article{tusing2000sounds,
  title={The sounds of dominance. Vocal precursors of perceived dominance during interpersonal influence},
  author={Tusing, Kyle James and Dillard, James Price},
  journal={Human Communication Research},
  volume={26},
  number={1},
  pages={148--171},
  year={2000},
  publisher={Wiley Online Library}
}

@article{choi2025acoustic,
  title={Acoustic-based Gender Differentiation in Speech-aware Language Models},
  author={Choi, Junhyuk and Seol, Jihwan and Kim, Nayeon and Cho, Chanhee and Cho, EunBin and Kim, Bugeun},
  journal={arXiv preprint arXiv:2509.21125},
  year={2025}
}

@misc{conti2025unheard,
      title={The Unheard Alternative: Contrastive Explanations for Speech-to-Text Models}, 
      author={Lina Conti and Dennis Fucci and Marco Gaido and Matteo Negri and Guillaume Wisniewski and Luisa Bentivogli},
      year={2025},
      eprint={2509.26543},
      archivePrefix={arXiv},
      primaryClass={cs.CL},
      url={https://arxiv.org/abs/2509.26543}, 
}

@inproceedings{sarti2023inseq,
  title={Inseq: An Interpretability Toolkit for Sequence Generation Models},
  author={Sarti, Gabriele and Feldhus, Nils and Sickert, Ludwig and Van Der Wal, Oskar},
  booktitle={Proceedings of the 61st Annual Meeting of the Association for Computational Linguistics (Volume 3: System Demonstrations)},
  pages={421--435},
  year={2023}
}

@inproceedings{stanovsky-etal-2019-evaluating,
    title = "Evaluating Gender Bias in Machine Translation",
    author = "Stanovsky, Gabriel  and
      Smith, Noah A.  and
      Zettlemoyer, Luke",
    editor = "Korhonen, Anna  and
      Traum, David  and
      M{\`a}rquez, Llu{\'i}s",
    booktitle = "Proceedings of the 57th Annual Meeting of the Association for Computational Linguistics",
    month = jul,
    year = "2019",
    address = "Florence, Italy",
    publisher = "Association for Computational Linguistics",
    url = "https://aclanthology.org/P19-1164/",
    doi = "10.18653/v1/P19-1164",
    pages = "1679--1684"
}

@inproceedings{escude-font-costa-jussa-2019-equalizing,
    title = "Equalizing Gender Bias in Neural Machine Translation with Word Embeddings Techniques",
    author = "Escud{\'e} Font, Joel  and
      Costa-juss{\`a}, Marta R.",
    editor = "Costa-juss{\`a}, Marta R.  and
      Hardmeier, Christian  and
      Radford, Will  and
      Webster, Kellie",
    booktitle = "Proceedings of the First Workshop on Gender Bias in Natural Language Processing",
    month = aug,
    year = "2019",
    address = "Florence, Italy",
    publisher = "Association for Computational Linguistics",
    url = "https://aclanthology.org/W19-3821/",
    doi = "10.18653/v1/W19-3821",
    pages = "147--154"
}

@inproceedings{manna-etal-2025-paying,
    title = "Are We Paying Attention to Her? Investigating Gender Disambiguation and Attention in Machine Translation",
    author = "Manna, Chiara  and
      Alishahi, Afra  and
      Blain, Fr{\'e}d{\'e}ric  and
      Vanmassenhove, Eva",
    editor = "Hackenbuchner, Jani{\c{c}}a  and
      Bentivogli, Luisa  and
      Daems, Joke  and
      Manna, Chiara  and
      Savoldi, Beatrice  and
      Vanmassenhove, Eva",
    booktitle = "Proceedings of the 3rd Workshop on Gender-Inclusive Translation Technologies (GITT 2025)",
    month = jun,
    year = "2025",
    address = "Geneva, Switzerland",
    publisher = "European Association for Machine Translation",
    url = "https://aclanthology.org/2025.gitt-1.1/",
    pages = "1--16",
    ISBN = "978-2-9701897-4-9"
}

@article{hillenbrand2009role,
  title={The role of f 0 and formant frequencies in distinguishing the voices of men and women},
  author={Hillenbrand, James M and Clark, Michael J},
  journal={Attention, Perception, \& Psychophysics},
  volume={71},
  number={5},
  pages={1150--1166},
  year={2009},
  publisher={Springer}
}

@article{mastromichalakis2025assumed,
  title={Assumed Identities: Quantifying Gender Bias in Machine Translation of Gender-Ambiguous Occupational Terms},
  author={Mastromichalakis, Orfeas Menis and Filandrianos, Giorgos and Symeonaki, Maria and Stamou, Giorgos},
  journal={arXiv preprint arXiv:2503.04372},
  year={2025}
}

@inproceedings{savoldi-etal-2024-harm,
    title = "What the Harm? Quantifying the Tangible Impact of Gender Bias in Machine Translation with a Human-centered Study",
    author = "Savoldi, Beatrice  and
      Papi, Sara  and
      Negri, Matteo  and
      Guerberof-Arenas, Ana  and
      Bentivogli, Luisa",
    editor = "Al-Onaizan, Yaser  and
      Bansal, Mohit  and
      Chen, Yun-Nung",
    booktitle = "Proceedings of the 2024 Conference on Empirical Methods in Natural Language Processing",
    month = nov,
    year = "2024",
    address = "Miami, Florida, USA",
    publisher = "Association for Computational Linguistics",
    url = "https://aclanthology.org/2024.emnlp-main.1002/",
    doi = "10.18653/v1/2024.emnlp-main.1002",
    pages = "18048--18076"
}

@inproceedings{papi2024good,
  title={When Good and Reproducible Results are a Giant with Feet of Clay: The Importance of Software Quality in NLP},
  author={Papi, Sara and Gaido, Marco and Pilzer, Andrea and Negri, Matteo},
  booktitle={Proceedings of the 62nd Annual Meeting of the Association for Computational Linguistics (Volume 1: Long Papers)},
  pages={3657--3672},
  year={2024}
}

@incollection{zimman-2020-transgender-lang,
    author = {Zimman, Lal},
    isbn = {9780190212926},
    title = {Transgender Language, Transgender Moment: Toward a Trans Linguistics},
    booktitle = {The Oxford Handbook of Language and Sexuality},
    publisher = {Oxford University Press},
    doi = {10.1093/oxfordhb/9780190212926.013.45},
    url = {https://doi.org/10.1093/oxfordhb/9780190212926.013.45},
    eprint = {https://academic.oup.com/book/0/chapter/358161844/chapter-ag-pdf/61662550/book\_42645\_section\_358161844.ag.pdf},
    year = {2020}
}

@article{savoldi2025mgente,
  title={mgente: A multilingual resource for gender-neutral language and translation},
  author={Savoldi, Beatrice and Cupin, Eleonora and Thind, Manjinder and Lauscher, Anne and Piergentili, Andrea and Negri, Matteo and Bentivogli, Luisa},
  journal={arXiv preprint arXiv:2501.09409},
  year={2025}
}

@inproceedings{piergentili2023hi,
  title={Hi Guys or Hi Folks? Benchmarking Gender-Neutral Machine Translation with the GeNTE Corpus},
  author={Piergentili, Andrea and Savoldi, Beatrice and Fucci, Dennis and Negri, Matteo and Bentivogli, Luisa},
  booktitle={Proceedings of the 2023 Conference on Empirical Methods in Natural Language Processing},
  pages={14124--14140},
  year={2023}
}

@article{Prates2018AssessingGBA,
  title={Assessing gender bias in machine translation: a case study with Google Translate},
  author={Marcelo O. R. Prates and Pedro H. C. Avelar and L. Lamb},
  journal={Neural Computing and Applications},
  year={2018},
  volume={32},
  pages={6363 - 6381},
  url={https://doi.org/10.1007/s00521-019-04144-6}
}

@book{butler1990feminism,
	author = {Judith Butler},
	publisher = {Routledge},
	title = {Gender Trouble: Feminism and the Subversion of Identity},
	year = {1990}
}

@article{chen-2025-airpods,
 author  = {Chen, Brian X.},
 year    = {2025},
 month   = {September},
 title   = {The New AirPods Can Translate Languages in Your Ears. This Is Profound.},
 journal = {The New York Times},
 url     = {https://www.nytimes.com/2025/09/18/technology/personaltech/new-airpods-language-translation-feature.html},
 urldate = {2025-18-09}
}

@inproceedings{curry-etal-2024-classist,
    title = "Classist Tools: Social Class Correlates with Performance in {NLP}",
    author = "Cercas Curry, Amanda  and
      Attanasio, Giuseppe  and
      Talat, Zeerak  and
      Hovy, Dirk",
    editor = "Ku, Lun-Wei  and
      Martins, Andre  and
      Srikumar, Vivek",
    booktitle = "Proceedings of the 62nd Annual Meeting of the Association for Computational Linguistics (Volume 1: Long Papers)",
    month = aug,
    year = "2024",
    address = "Bangkok, Thailand",
    publisher = "Association for Computational Linguistics",
    url = "https://aclanthology.org/2024.acl-long.682/",
    doi = "10.18653/v1/2024.acl-long.682",
    pages = "12643--12655"
}

@article{savoldi2025decade,
  title={A decade of gender bias in machine translation},
  author={Savoldi, Beatrice and Bastings, Jasmijn and Bentivogli, Luisa and Vanmassenhove, Eva},
  journal={Patterns},
  volume={6},
  number={6},
  year={2025},
  publisher={Elsevier}
}

@article{labov1964phonological,
  title={Phonological correlates of social stratification},
  author={Labov, William},
  journal={American Anthropologist},
  volume={66},
  number={6},
  pages={164--176},
  year={1964},
  publisher={JSTOR}
}

@inproceedings{wang2020fairseq,
  title={Fairseq S2T: Fast Speech-to-Text Modeling with Fairseq},
  author={Wang, Changhan and Tang, Yun and Ma, Xutai and Wu, Anne and Okhonko, Dmytro and Pino, Juan},
  booktitle={Proceedings of the 1st Conference of the Asia-Pacific Chapter of the Association for Computational Linguistics and the 10th International Joint Conference on Natural Language Processing: System Demonstrations},
  pages={33--39},
  year={2020}
}

@article{simpson2009phonetic,
  title={Phonetic differences between male and female speech},
  author={Simpson, Adrian P},
  journal={Language and linguistics compass},
  volume={3},
  number={2},
  pages={621--640},
  year={2009},
  publisher={Wiley Online Library}
}

@article{choi2025voicebbq,
  title={VoiceBBQ: Investigating Effect of Content and Acoustics in Social Bias of Spoken Language Model},
  author={Choi, Junhyuk and Oh, Ro-hoon and Seol, Jihwan and Kim, Bugeun},
  journal={arXiv preprint arXiv:2509.21108},
  year={2025}
}

@inproceedings{lin2024social,
  title={On the social bias of speech self-supervised models},
  author={Lin, Yi-Cheng and Lin, Tzu-Quan and Lin, Hsi-Che and Liu, Andy T and Lee, Hung-yi},
  booktitle={Proc. Interspeech 2024},
  pages={4638--4642},
  year={2024}
}

@article{feng2021quantifying,
  title={Quantifying bias in automatic speech recognition},
  author={Feng, Siyuan and Kudina, Olya and Halpern, Bence Mark and Scharenborg, Odette},
  journal={arXiv preprint arXiv:2103.15122},
  year={2021}
}

@inproceedings{zanon2022study,
  title={A Study of Gender Impact in Self-supervised Models for Speech-to-Text Systems},
  author={Zanon Boito, Marcely and Besacier, Laurent and Tomashenko, Natalia and Est{\`e}ve, Yannick},
  booktitle={Proc. Interspeech 2022},
  pages={1278--1282},
  year={2022}
}

@INPROCEEDINGS{chien2024balancing,
  author={Chien, Woan-Shiuan and Upadhyay, Shreya G. and Lee, Chi-Chun},
  booktitle={ICASSP 2024 - 2024 IEEE International Conference on Acoustics, Speech and Signal Processing (ICASSP)}, 
  title={Balancing Speaker-Rater Fairness for Gender-Neutral Speech Emotion Recognition}, 
  year={2024},
  volume={},
  number={},
  pages={11861-11865},
  doi={10.1109/ICASSP48485.2024.10447167}}

@inproceedings{slaughter-etal-2023-pre,
    title = "Pre-trained Speech Processing Models Contain Human-Like Biases that Propagate to Speech Emotion Recognition",
    author = "Slaughter, Isaac  and
      Greenberg, Craig  and
      Schwartz, Reva  and
      Caliskan, Aylin",
    editor = "Bouamor, Houda  and
      Pino, Juan  and
      Bali, Kalika",
    booktitle = "Findings of the Association for Computational Linguistics: EMNLP 2023",
    month = dec,
    year = "2023",
    address = "Singapore",
    publisher = "Association for Computational Linguistics",
    url = "https://aclanthology.org/2023.findings-emnlp.602/",
    doi = "10.18653/v1/2023.findings-emnlp.602",
    pages = "8967--8989"
}

@inproceedings{lin2024emo,
  title={Emo-bias: A Large Scale Evaluation of Social Bias on Speech Emotion Recognition},
  author={Lin, Yi-Cheng and Wu, Haibin and Chou, Huang-Cheng and Lee, Chi-Chun and Lee, Hung-yi},
  booktitle={Proc. Interspeech 2024},
  pages={4633--4637},
  year={2024}
}

@inproceedings{meng2022don,
  title={Don't speak too fast: The impact of data bias on self-supervised speech models},
  author={Meng, Yen and Chou, Yi-Hui and Liu, Andy T and Lee, Hung-yi},
  booktitle={ICASSP 2022-2022 IEEE International Conference on Acoustics, Speech and Signal Processing (ICASSP)},
  pages={3258--3262},
  year={2022},
  organization={IEEE}
}

@inproceedings{lin2024listen,
  title={Listen and speak fairly: a study on semantic gender bias in speech integrated large language models},
  author={Lin, Yi-Cheng and Lin, Tzu-Quan and Yang, Chih-Kai and Lu, Ke-Han and Chen, Wei-Chih and Kuan, Chun-Yi and Lee, Hung-yi},
  booktitle={2024 IEEE Spoken Language Technology Workshop (SLT)},
  pages={439--446},
  year={2024},
  organization={IEEE}
}

@article{trinh2020directly,
  title={Directly comparing the listening strategies of humans and machines},
  author={Trinh, Viet Anh and Mandel, Michael},
  journal={IEEE/ACM Transactions on Audio, Speech, and Language Processing},
  volume={29},
  pages={312--323},
  year={2020},
  publisher={IEEE}
}

@article{markertvisualizing,
  title={Visualizing Automatic Speech Recognition--Means for a Better Understanding?},
  author={Markert, Karla and Parracone, Romain and Kulakov, Mykhailo and Sperl, Philip and Kao, Ching-Yu and B{\"o}ttinger, Konstantin},
  journal={ ISCA Symposium on Security and Privacy in Speech Communication},
  year={2021}
}

@INPROCEEDINGS{wu2023explanations,
  author={Wu, Xiaoliang and Bell, Peter and Rajan, Ajitha},
  booktitle={ICASSP 2023 - 2023 IEEE International Conference on Acoustics, Speech and Signal Processing (ICASSP)}, 
  title={Explanations for Automatic Speech Recognition}, 
  year={2023},
  volume={},
  number={},
  pages={1-5},
  doi={10.1109/ICASSP49357.2023.10094635}}

@inproceedings{wu2024can,
  title={Can we trust explainable ai methods on asr? an evaluation on phoneme recognition},
  author={Wu, Xiaoliang and Bell, Peter and Rajan, Ajitha},
  booktitle={ICASSP 2024-2024 IEEE International Conference on Acoustics, Speech and Signal Processing (ICASSP)},
  pages={10296--10300},
  year={2024},
  organization={IEEE}
}

@inproceedings{mohebbi2023homophone,
  title={Homophone Disambiguation Reveals Patterns of Context Mixing in Speech Transformers},
  author={Mohebbi, Hosein and Chrupa{\l}a, Grzegorz and Zuidema, Willem and Alishahi, Afra},
  booktitle={Proceedings of the 2023 Conference on Empirical Methods in Natural Language Processing},
  pages={8249--8260},
  year={2023}
}

@inproceedings{Saunders2020ReducingGBA,
  title={Reducing Gender Bias in Neural Machine Translation as a Domain Adaptation Problem},
  author={Danielle Saunders and B. Byrne},
  booktitle={Annual Meeting of the Association for Computational Linguistics},
  year={2020},
  url={https://www.aclweb.org/anthology/2020.acl-main.690.pdf}
}

@inproceedings{vanmassenhove2018getting,
  title={Getting Gender Right in Neural Machine Translation},
  author={Vanmassenhove, Eva and Hardmeier, Christian and Way, Andy},
  booktitle={Proceedings of the 2018 Conference on Empirical Methods in Natural Language Processing},
  pages={3003--3008},
  year={2018}
}

@inproceedings{saunders2022first,
  title={First the Worst: Finding Better Gender Translations During Beam Search},
  author={Saunders, Danielle and Sallis, Rosie and Byrne, Bill},
  booktitle={Findings of the Association for Computational Linguistics: ACL 2022},
  pages={3814--3823},
  year={2022}
}

@inproceedings{calado-2025-myths,
    title = "Some Myths About Bias: A Queer Studies Reading Of Gender Bias In {NLP}",
    author = "Calado, Filipa",
    editor = "Fale{\'n}ska, Agnieszka  and
      Basta, Christine  and
      Costa-juss{\`a}, Marta  and
      Sta{\'n}czak, Karolina  and
      Nozza, Debora",
    booktitle = "Proceedings of the 6th Workshop on Gender Bias in Natural Language Processing (GeBNLP)",
    month = aug,
    year = "2025",
    address = "Vienna, Austria",
    publisher = "Association for Computational Linguistics",
    url = "https://aclanthology.org/2025.gebnlp-1.29/",
    doi = "10.18653/v1/2025.gebnlp-1.29",
    pages = "338--346",
    ISBN = "979-8-89176-277-0"
}

@inproceedings{vanmassenhove2019lost,
  title={Lost in Translation: Loss and Decay of Linguistic Richness in Machine Translation},
  author={Vanmassenhove, Eva and Shterionov, Dimitar and Way, Andy},
  booktitle={Proceedings of Machine Translation Summit XVII: Research Track},
  pages={222--232},
  year={2019}
}

@inproceedings{wisniewski2022analyzing,
  title={Analyzing gender translation errors to identify information flows between the encoder and decoder of a NMT system},
  author={Wisniewski, Guillaume and Zhu, Lichao and Ballier, Nicolas and Yvon, Fran{\c{c}}ois},
  booktitle={Proceedings of the Fifth BlackboxNLP Workshop on Analyzing and Interpreting Neural Networks for NLP},
  pages={153--163},
  year={2022}
}

@article{kraus2017signs,
  title={Signs of social class: The experience of economic inequality in everyday life},
  author={Kraus, Michael W and Park, Jun Won and Tan, Jacinth JX},
  journal={Perspectives on Psychological Science},
  volume={12},
  number={3},
  pages={422--435},
  year={2017},
  publisher={Sage Publications Sage CA: Los Angeles, CA}
}

@article{Thomas2010TeachingAL,
  title={Teaching and Learning Guide for: Phonological and Phonetic Characteristics of African American Vernacular English},
  author={Thomas, Erik R.},
  journal={Lang. Linguistics Compass},
  year={2010},
  volume={4},
  pages={737-741},
  url={https://api.semanticscholar.org/CorpusID:12831627}
}

@article{azul2015varied,
  title={On the varied and complex factors affecting gender diverse people's vocal situations: Implications for clinical practice},
  author={Azul, David},
  journal={Perspectives on Voice and Voice Disorders},
  volume={25},
  number={2},
  pages={75--86},
  year={2015},
  publisher={American Speech-Language-Hearing Association}
}

@book{kreiman2011foundations,
  title={Foundations of voice studies: An interdisciplinary approach to voice production and perception},
  author={Kreiman, Jody and Sidtis, Diana},
  year={2011},
  publisher={John Wiley \& Sons}
}

@inproceedings{sawalha2013effects,
  title={The effects of speakers' gender, age, and region on overall performance of Arabic automatic speech recognition systems using the phonetically rich and balanced Modern Standard Arabic speech corpus},
  author={Sawalha, Majdi and Abu Shariah, Mohammad},
  booktitle={Proceedings of the 2nd Workshop of Arabic Corpus Linguistics WACL-2},
  year={2013},
  organization={Leeds}
}

@inproceedings{liu2022towards,
  title={Towards measuring fairness in speech recognition: Casual conversations dataset transcriptions},
  author={Liu, Chunxi and Picheny, Michael and Sar{\i}, Leda and Chitkara, Pooja and Xiao, Alex and Zhang, Xiaohui and Chou, Mark and Alvarado, Andres and Hazirbas, Caner and Saraf, Yatharth},
  booktitle={ICASSP 2022-2022 IEEE International Conference on Acoustics, Speech and Signal Processing (ICASSP)},
  pages={6162--6166},
  year={2022},
  organization={IEEE}
}

@inproceedings{rajan2022aequevox,
  title={Aequevox: Automated fairness testing of speech recognition systems},
  author={Rajan, Sai Sathiesh and Udeshi, Sakshi and Chattopadhyay, Sudipta},
  booktitle={International Conference on Fundamental Approaches to Software Engineering},
  pages={245--267},
  year={2022},
  organization={Springer International Publishing Cham}
}

@inproceedings{gaido-etal-2020-breeding,
    title = "Breeding Gender-aware Direct Speech Translation Systems",
    author = "Gaido, Marco  and
      Savoldi, Beatrice  and
      Bentivogli, Luisa  and
      Negri, Matteo  and
      Turchi, Marco",
    editor = "Scott, Donia  and
      Bel, Nuria  and
      Zong, Chengqing",
    booktitle = "Proceedings of the 28th International Conference on Computational Linguistics",
    month = dec,
    year = "2020",
    address = "Barcelona, Spain (Online)",
    publisher = "International Committee on Computational Linguistics",
    url = "https://aclanthology.org/2020.coling-main.350/",
    doi = "10.18653/v1/2020.coling-main.350",
    pages = "3951--3964"
}

@inproceedings{adda2005speech,
  title={Do speech recognizers prefer female speakers?},
  author={Adda-Decker, Martine and Lamel, Lori},
  booktitle={Interspeech},
  pages={2205--2208},
  year={2005}
}

@inproceedings{savoldi2022dynamics,
  title={On the dynamics of gender learning in speech translation},
  author={Savoldi, Beatrice and Gaido, Marco and Bentivogli, Luisa and Negri, Matteo and Turchi, Marco},
  booktitle={Proceedings of the 4th Workshop on Gender Bias in Natural Language Processing (GeBNLP)},
  pages={94--111},
  year={2022},
  organization={Association for Computational Linguistics}
}

@INPROCEEDINGS{lin2024spokenstereoset,

  author={Lin, Yi-Cheng and Chen, Wei-Chih and Lee, Hung-Yi},

  booktitle={2024 IEEE Spoken Language Technology Workshop (SLT)}, 

  title={Spoken Stereoset: on Evaluating Social Bias Toward Speaker in Speech Large Language Models}, 

  year={2024},

  volume={},

  number={},

  pages={871-878},

  doi={10.1109/SLT61566.2024.10832259}}

@article{chowdhury2024end,
  title={What do end-to-end speech models learn about speaker, language and channel information? a layer-wise and neuron-level analysis},
  author={Chowdhury, Shammur Absar and Durrani, Nadir and Ali, Ahmed},
  journal={Computer Speech \& Language},
  volume={83},
  pages={101539},
  year={2024},
  publisher={Elsevier}
}

@inproceedings{de2022probing,
  title={Probing phoneme, language and speaker information in unsupervised speech representations},
  author={de Seyssel, Maureen and Lavechin, Marvin and Adi, Yossi and Dupoux, Emmanuel and Wisniewski, Guillaume},
  booktitle={Interspeech 2022-23rd INTERSPEECH Conference},
  year={2022}
}

@inproceedings{guillaume2024gender,
  title={Gender and language identification in Multilingual Models of Speech: exploring the genericity and robustness of speech representations},
  author={Guillaume, S{\'e}verine and Fily, Maxime and Michaud, Alexis and Wisniewski, Guillaume},
  booktitle={Interspeech 2024},
  pages={3330--3334},
  year={2024},
  organization={ISCA}
}

@inproceedings{krishnan2024encoding,
  title={On the Encoding of Gender in Transformer-based ASR Representations},
  author={Krishnan, Aravind and Abdullah, Badr M and Klakow, Dietrich},
  booktitle={Proc. Interspeech 2024},
  pages={3090--3094},
  year={2024}
}

@article{seamless2025joint,
  title={Joint speech and text machine translation for up to 100 languages},
  journal={Nature},
  author={Barrault, Lo{\"\i}c and Chung, Yu-An and Meglioli, Mariano Cora and Dale, David and Dong, Ning and Duquenne, Paul-Ambroise and Elsahar, Hady and Gong, Hongyu and Heffernan, Kevin and Hoffman, John and others},
  volume={637},
  number={8046},
  pages={587--593},
  year={2025},
  publisher={Nature Publishing Group UK London}
}

@article{bansal2025addressing,
  title={Addressing speaker gender bias in large scale speech translation systems},
  author={Bansal, Shubham and Joshi, Vikas and Chadha, Harveen and Mehta, Rupeshkumar and Li, Jinyu},
  journal={arXiv preprint arXiv:2501.05989},
  year={2025}
}

@inproceedings{zeineldeen2021investigating,
  title     = {Investigating Methods to Improve Language Model Integration for Attention-Based Encoder-Decoder ASR Models},
  author    = {Mohammad Zeineldeen and Aleksandr Glushko and Wilfried Michel and Albert Zeyer and Ralf Schlüter and Hermann Ney},
  year      = {2021},
  booktitle = {Interspeech 2021},
  pages     = {2856--2860},
  doi       = {10.21437/Interspeech.2021-1255},
  issn      = {2958-1796},
}

@inproceedings{fucci2023integrating,
  title={Integrating Language Models into Direct Speech Translation: An Inference-Time Solution to Control Gender Inflection},
  author={Fucci, Dennis and Gaido, Marco and Papi, Sara and Cettolo, Mauro and Negri, Matteo and Bentivogli, Luisa},
  booktitle={Proceedings of the 2023 Conference on Empirical Methods in Natural Language Processing},
  pages={11505--11517},
  year={2023}
}

@inproceedings{variani2020hybrid,
  title={Hybrid autoregressive transducer (hat)},
  author={Variani, Ehsan and Rybach, David and Allauzen, Cyril and Riley, Michael},
  booktitle={ICASSP 2020-2020 IEEE International Conference on Acoustics, Speech and Signal Processing (ICASSP)},
  pages={6139--6143},
  year={2020},
  organization={IEEE}
}

@inproceedings{meng2021internal,
  title={Internal language model estimation for domain-adaptive end-to-end speech recognition},
  author={Meng, Zhong and Parthasarathy, Sarangarajan and Sun, Eric and Gaur, Yashesh and Kanda, Naoyuki and Lu, Liang and Chen, Xie and Zhao, Rui and Li, Jinyu and Gong, Yifan},
  booktitle={2021 IEEE Spoken Language Technology Workshop (SLT)},
  pages={243--250},
  year={2021},
  organization={IEEE}
}

@inproceedings{conti2023using,
  title={Using Artificial French Data to Understand the Emergence of Gender Bias in Transformer Language Models},
  author={Conti, Lina and Wisniewski, Guillaume},
  booktitle={Proceedings of the 2023 Conference on Empirical Methods in Natural Language Processing},
  pages={10362--10371},
  year={2023}
}

@inproceedings{piergentili2024neogate,
  title={Enhancing Gender-Inclusive Machine Translation with Neomorphemes and Large Language Models},
  author={Piergentili, Andrea and Savoldi, Beatrice and Negri, Matteo and Bentivogli, Luisa},
  booktitle={Proceedings of the 25th Annual Conference of the European Association for Machine Translation (Volume 1)},
  pages={300--314},
  year={2024}
}

@inproceedings{tang2024salmonn,
  title={SALMONN: Towards Generic Hearing Abilities for Large Language Models},
  author={Tang, Changli and Yu, Wenyi and Sun, Guangzhi and Chen, Xianzhao and Tan, Tian and Li, Wei and Lu, Lu and MA, Zejun and Zhang, Chao},
  booktitle={The Twelfth International Conference on Learning Representations},
  year={2024}
}

@inproceedings{wu2023decoder,
  title={On decoder-only architecture for speech-to-text and large language model integration},
  author={Wu, Jian and Gaur, Yashesh and Chen, Zhuo and Zhou, Long and Zhu, Yimeng and Wang, Tianrui and Li, Jinyu and Liu, Shujie and Ren, Bo and Liu, Linquan and others},
  booktitle={2023 IEEE Automatic Speech Recognition and Understanding Workshop (ASRU)},
  pages={1--8},
  year={2023},
  organization={IEEE}
}

@article{rubenstein2023audiopalm,
  title={Audiopalm: A large language model that can speak and listen},
  author={Rubenstein, Paul K and Asawaroengchai, Chulayuth and Nguyen, Duc Dung and Bapna, Ankur and Borsos, Zal{\'a}n and Quitry, F{\'e}lix de Chaumont and Chen, Peter and Badawy, Dalia El and Han, Wei and Kharitonov, Eugene and others},
  journal={arXiv preprint arXiv:2306.12925},
  year={2023}
}

@inproceedings{xu2023recent,
  title={Recent advances in direct speech-to-text translation},
  author={Xu, Chen and Ye, Rong and Dong, Qianqian and Zhao, Chengqi and Ko, Tom and Wang, Mingxuan and Xiao, Tong and Zhu, Jingbo},
  booktitle={Proceedings of the Thirty-Second International Joint Conference on Artificial Intelligence},
  pages={6796--6804},
  year={2023}
}

@article{yang2025towards,
  title={Towards holistic evaluation of large audio-language models: A comprehensive survey},
  author={Yang, Chih-Kai and Ho, Neo S and Lee, Hung-yi},
  journal={arXiv preprint arXiv:2505.15957},
  year={2025}
}

\section{\lc{Appendices}}

\subsection{Model Gender Accuracy}
\label{app:acc}

Tables~\ref{tab:gender-accuracies-transformer} and \ref{tab:gender-accuracies-conformer} report gender accuracy for all models and language pairs, complementing the model descriptions in Section~\ref{sec:exp-setup}. By gender accuracy, we mean the proportion of correct gender realizations among terms where the model generates one of the MuST-SHE annotated forms \cite{gaido-etal-2020-breeding}.

\begin{table}[ht]
\centering
\begin{tabular}{|l|c|c|c|}
\hline
 & es & fr & it \\
\hline
Feminine  & 80.60\,\% & 77.35\,\% & 77.09\,\% \\
Masculine & 91.41\,\% & 91.85\,\% & 94.40\,\% \\
\hline
\end{tabular}
\caption{Gender accuracy for the Transformer model \cite{wang2020fairseq}.}
\label{tab:gender-accuracies-transformer}
\end{table}

\begin{table}[ht]
\centering
\begin{tabular}{|l|c|c|c|}
\hline
 & es & fr & it \\
\hline
Feminine  & 39.21\,\% & 49.80\,\% & 46.38\,\% \\
Masculine & 76.74\,\% & 72.50\,\% & 75.76\,\% \\
\hline
\end{tabular}
\caption{Gender accuracy for the Conformer models \cite{papi2024good}.}
\label{tab:gender-accuracies-conformer}
\end{table}

\subsection{Training Data Analysis}
\label{app:training-data}

Table~\ref{tab:training-data-conf} extends the results of Table~\ref{tab:training-data} to the Conformer models.

\begin{table}[!ht]
\begin{center}
\begin{subtable}{0.32\textwidth}
\centering
\begin{tabular}{|c|c|c|}
\cline{2-3}
\multicolumn{1}{c|}{} & More Freq. & Less Freq. \\
\hline
F & 22 & 116 \\
M & 284 & 31 \\
\hline
\end{tabular}
\caption{Spanish}
\end{subtable}
\hfill
\begin{subtable}{0.32\textwidth}
\centering
\begin{tabular}{|c|c|c|}
\cline{2-3}
\multicolumn{1}{c|}{} & More Freq. & Less Freq. \\
\hline
F & 19 & 131 \\
M & 213 & 16 \\
\hline
\end{tabular}
\caption{French}
\end{subtable}
\hfill
\begin{subtable}{0.32\textwidth}
\centering
\begin{tabular}{|c|c|c|}
\cline{2-3}
\multicolumn{1}{c|}{} & More Freq. & Less Freq. \\
\hline
F & 13 & 105 \\
M & 179 & 12 \\
\hline
\end{tabular}
\caption{Italian}
\end{subtable}
\caption{Distribution of examples by predicted gender and whether the predicted gender is more or less prevalent in the training data for that term. Results for the Conformer models \cite{papi2024good}.}
\label{tab:training-data-conf}
\end{center}
\end{table}

\subsection{Internal Language Model Analysis}
\label{app:ilm}

Tables~\ref{tab:ilm-transformer}, \ref{tab:ilm-conformer}, and \ref{tab:ilm-conf} extend the results of Section~\ref{sec:ilm} to all languages and models.

\begin{table}[ht]
\centering
\begin{tabular}{|l|c|c|c|}
\hline
 & es & fr & it \\
\hline
All terms & 0.76 & 0.81 & 0.73 \\
Fem.\ predictions & 0.65 & 0.71 & 0.58 \\
Masc.\ predictions & 0.85 & 0.88 & 0.85 \\
\hline
\end{tabular}
\caption{ILM masculine preference for the Transformer model \cite{wang2020fairseq}.}
\label{tab:ilm-transformer}
\end{table}

\begin{table}[ht]
\centering
\begin{tabular}{|l|c|c|c|}
\hline
 & es & fr & it \\
\hline
All terms & 0.64 & 0.63 & 0.64 \\
Fem.\ predictions & 0.49 & 0.48 & 0.49 \\
Masc.\ predictions & 0.71 & 0.73 & 0.74 \\
\hline
\end{tabular}
\caption{ILM masculine preference for the Conformer models \cite{papi2024good}.}
\label{tab:ilm-conformer}
\end{table}

\begin{table}[!ht]
\begin{center}
\begin{subtable}{0.32\textwidth}
\centering
\begin{tabular}{|c|c|c|}
\cline{2-3}
\multicolumn{1}{c|}{} & Higher Prob. & Lower Prob. \\
\hline
F & 72 & 66 \\
M & 228 & 87 \\
\hline
\end{tabular}
\caption{Spanish}
\end{subtable}
\hfill
\begin{subtable}{0.32\textwidth}
\centering
\begin{tabular}{|c|c|c|}
\cline{2-3}
\multicolumn{1}{c|}{} & Higher Prob. & Lower Prob. \\
\hline
F & 72 & 78 \\
M & 174 & 55 \\
\hline
\end{tabular}
\caption{French}
\end{subtable}
\hfill
\begin{subtable}{0.32\textwidth}
\centering
\begin{tabular}{|c|c|c|}
\cline{2-3}
\multicolumn{1}{c|}{} & Higher Prob. & Lower Prob. \\
\hline
F & 56 & 62 \\
M & 148 & 43 \\
\hline
\end{tabular}
\caption{Italian}
\end{subtable}
\caption{Distribution of examples by predicted gender and whether the ILM assigns higher or lower probability to the alternative. Results for the Conformer models \cite{papi2024good}.}
\label{tab:ilm-conf}
\end{center}
\end{table}

\subsection{Feature Attribution}
\label{app:feat-att}

\begin{figure*}[ht]
    \centering
    \begin{subfigure}[b]{0.48\linewidth}
        \includegraphics[width=\linewidth]{frequency_scores.png}
        \caption{Transformer, en→it}
    \end{subfigure}
    \hfill
    \begin{subfigure}[b]{0.48\linewidth}
        \includegraphics[width=\linewidth]{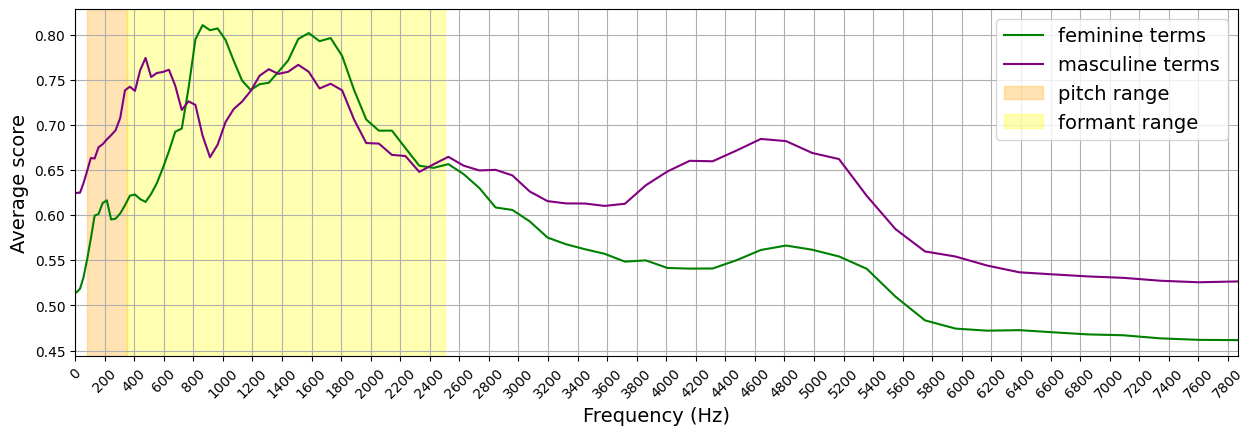}
        \caption{Transformer, en→es}
    \end{subfigure}
    \vspace{0.5em}
    \begin{subfigure}[b]{0.48\linewidth}
        \includegraphics[width=\linewidth]{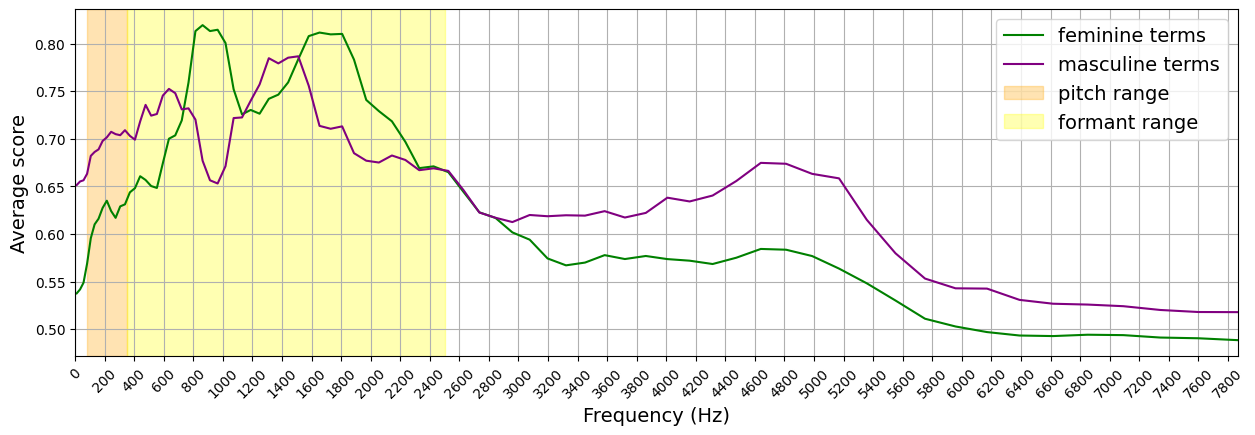}
        \caption{Transformer, en→fr}
    \end{subfigure}
    \hfill
    \begin{subfigure}[b]{0.48\linewidth}
        \includegraphics[width=\linewidth]{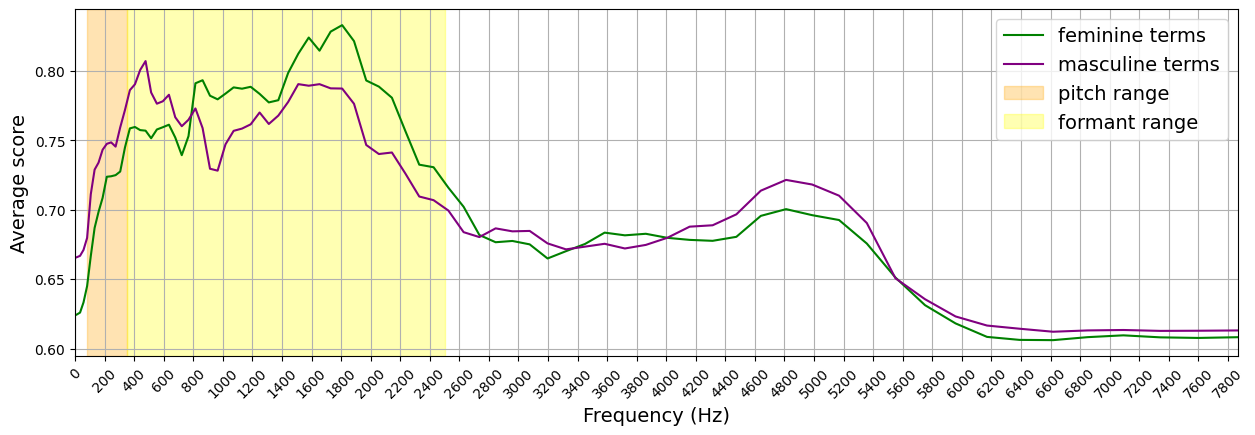}
        \caption{Conformer, en→it}
    \end{subfigure}
    \vspace{0.5em}
    \begin{subfigure}[b]{0.48\linewidth}
        \includegraphics[width=\linewidth]{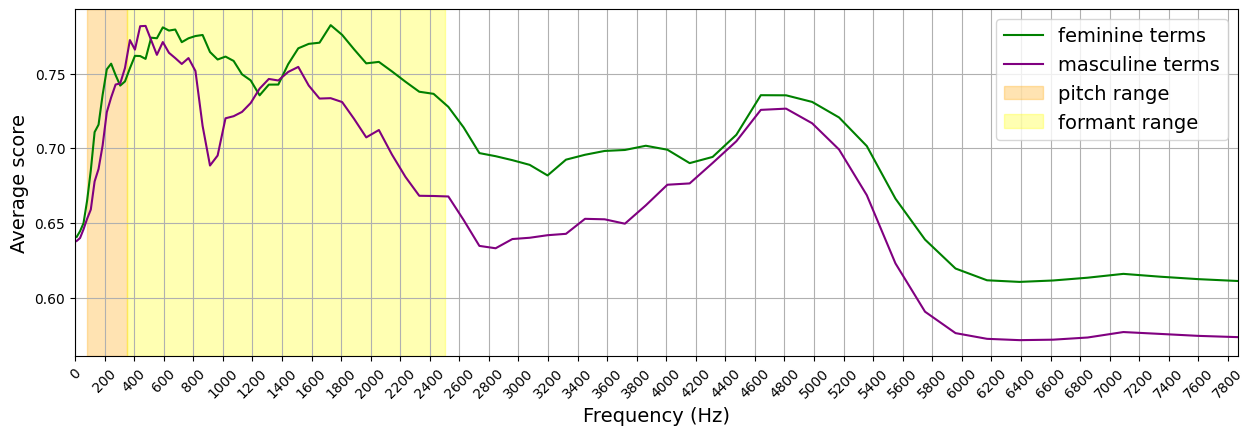}
        \caption{Conformer, en→es}
    \end{subfigure}
    \hfill
    \begin{subfigure}[b]{0.48\linewidth}
        \includegraphics[width=\linewidth]{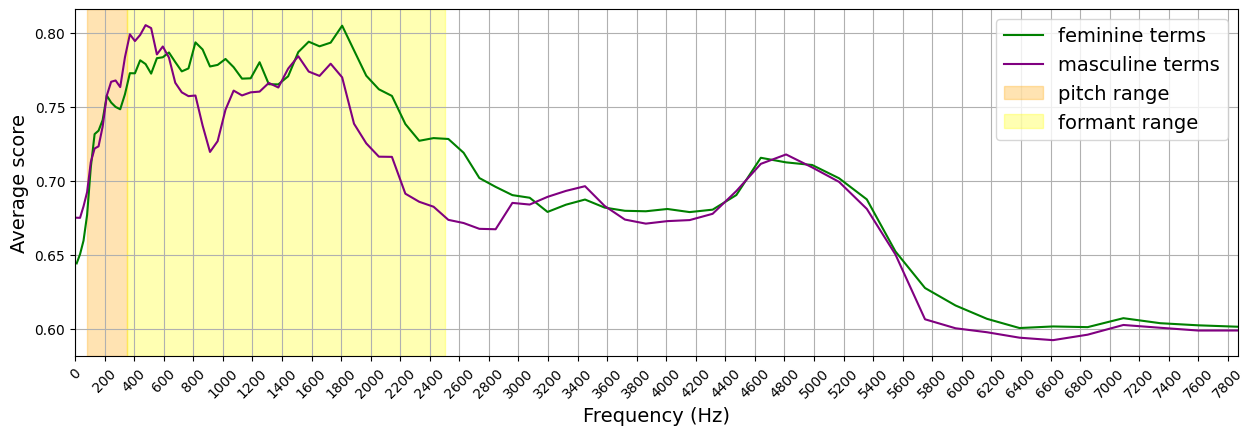}
        \caption{Conformer, en→fr}
    \end{subfigure}
    \caption{Average saliency scores per frequency bin (max-pooled over the time dimension, then averaged across all gender terms that flip), for the Transformer \cite{wang2020fairseq} and Conformer \cite{papi2024good} models across all three language pairs. Results are shown separately for feminine and masculine terms. Shaded regions mark the pitch range (80--350~Hz) and formant range (350--2500~Hz).}
    \label{fig:frequency-scores-full}
\end{figure*}

Tables~\ref{tab:self-ref-percent-conf} and \ref{tab:top-words-conformer} extend the results of Section~\ref{sec:feat-attribution} to the Conformer models, and Figure~\ref{fig:frequency-scores-full} extends the results in Figure~\ref{fig:frequency-scores} to all languages and models.

\begin{table}[!ht]
    \centering
    \begin{tabular}{c|ccc|ccc|}
        \cline{2-7}
        & \multicolumn{3}{c|}{`I'} & \multicolumn{3}{c|}{Self-referential} \\
        \cline{2-7}
         & es & fr & it & es & fr & it \\
         \hline
       \multicolumn{1}{|c|}{Flip} & \lc{12.4} & \lc{11.8} & \lc{16.5} & \lc{18.3} & \lc{14.7} & \lc{19.5} \\
       \multicolumn{1}{|c|}{All} & 10.1 & 9.0 & 15.5 & 14.3 & 12.1 & \lc{20.1} \\
        \hline
    \end{tabular}
    \caption{Percentage of examples where the top-scoring source word is “I” or a self-referential word, for examples that flip and for all examples. Results for the Conformer models \cite{papi2024good}.}
    \label{tab:self-ref-percent-conf}
\end{table}

\begin{table}[!ht]
\begin{center}
\begin{tabularx}{\columnwidth}{|lX|lX|lX|}
\hline
\multicolumn{2}{|c|}{es} & \multicolumn{2}{c}{fr} & \multicolumn{2}{|c|}{it} \\
\hline
I & 17 & I & 12 & I & 18 \\
as & 9 & by & 7 & sure & 3 \\
I'm & 5 & was & 6 & and & 3 \\
grateful & 4 & love & 4 & went & 2 \\
engineer & 4 & became & 3 & the & 2 \\
of & 4 & convinced & 3 & educator & 2 \\
was & 3 & me & 3 & I've & 2 \\
tired & 3 & came & 3 & scientist & 2 \\
sure & 3 & published & 2 & curator & 2 \\
myself & 3 & you're & 2 & when & 2 \\
\hline
\end{tabularx}
\caption{Most frequent top-scoring source words for examples that flip, with the number of examples for which they receive the highest saliency score in the source sentence. Results for the Conformer models \cite{papi2024good}.}
\label{tab:top-words-conformer}
\end{center}
\end{table}

\end{document}